\documentclass[10pt,twocolumn,letterpaper]{article}
\usepackage{multirow} 
\usepackage{algorithm}
\usepackage{algorithmic}
\usepackage{url}
\usepackage{pgfplots}
\usepackage{placeins}
\usepackage{bm}  
\pgfplotsset{compat=1.18}
\usepgfplotslibrary{groupplots}
\usepackage{filecontents}    

\usepackage{amsmath}
\usepackage{amssymb}
\usepackage{amsfonts}

\usepackage{booktabs}

\usepackage{subcaption}

\newcommand{\etal}{\textit{et al}.}

\definecolor{cvprblue}{rgb}{0.21,0.49,0.74}
\usepackage[pagebackref,breaklinks,colorlinks,allcolors=cvprblue]{hyperref}
\usepackage[nameinlink,capitalise]{cleveref}
\title{Frontend Response-Oriented Input Transformation for Transferable Adversarial Attacks}
\author{Quan Liu\\
\and
 Feng Ye\\
\and
 Chenhao Lu\\
 \and
 Shuming Zhen\\
 \and
 Guanliang Huang\\
 \and
 Lunzhe Chen\\
 \and
 Xudong Ke
}

\begin{document}
\maketitle
\begin{abstract}
Input transformation-based attacks improve adversarial transferability by aggregating gradients over transformed inputs. Existing analyses mainly explain their efficacy from image diversity, semantic preservation, attention variance or hypothesis space augmentation, yet overlook the critical role of model frontend responses. 
 In this paper, we revisit transformation-based attacks from an implicit ensemble perspective: each transformation can be viewed as a pre-processing operator before the surrogate model, inducing a distinct frontend response for gradient aggregation. Based on this view, we propose FRO, a Frontend Response-Oriented input transformation method that enriches such responses through two complementary operators. The Local Scaling Operator perturbs local content sampling via block-wise stretch-and-shrink operations, while the Projection Operator modifies global spatial organization through coherent perspective deformation. Together, they produce structured transformed views to optimize transferable adversarial perturbations. 
Experiments on an ImageNet subset show that FRO consistently improves black-box transferability across diverse CNN and Vision Transformer models. We further analyze the effect of implicit ensemble size and evaluate different transformation-based methods under a unified ensemble scale, demonstrating the superiority of designing input transformations from the perspective of front-end response ensembles.
\end{abstract}    
\section{Introduction}

Deep Neural Networks (DNNs) have achieved remarkable performance in visual tasks such as image classification \cite{Krizhevsky2012ImageNetCW}, object detection \cite{Redmon2015YouOL}, and semantic segmentation \cite{Shelhamer2014FullyCN}. However, DNNs remain vulnerable to adversarial examples \cite{Goodfellow2014ExplainingAH,Kurakin2016AdversarialEI}, where small and often imperceptible perturbations can mislead well-trained models. This vulnerability raises security concerns in safety-critical applications, including autonomous driving \cite{Yang2020FindingPA}, medical image analysis \cite{Ma2019UnderstandingAA}, and identity verification \cite{Sharif2016AccessorizeTA}. Studying adversarial examples is therefore important for evaluating and improving model robustness.

Adversarial attacks are broadly divided into white-box and black-box attacks. In white-box settings, attackers have full access to the target model and can directly optimize perturbations using model gradients. Representative methods include FGSM \cite{Goodfellow2014ExplainingAH}, MI-FGSM \cite{Dong2017DiscoveringAE}, and C\&W \cite{Carlini2016TowardsET}. In black-box settings, attackers have limited or no access to the target model, which better reflects practical attack scenarios \cite{Papernot2015TheLO,Wang2021DIAAAI}. A common black-box strategy is to exploit adversarial transferability, whereby adversarial examples crafted on an accessible source model can also fool unseen target models \cite{Liu2016DelvingIT,Madry2017TowardsDL}. Since standard white-box attacks often exhibit limited transferability \cite{Papernot2016PracticalBA}, improving cross-model transferability remains an important problem.

Following the taxonomy of a recent comprehensive survey \cite{wang2026devling}, existing transfer-based attacks can be broadly categorized into gradient-based attacks \cite{Li2024FoolmixST}, input transformation-based attacks \cite{Wang2023TransferAttack,11094976}, advanced-objective-function attacks \cite{pmlr-v267-liu25cd}, generation-based attacks \cite{10716799}, model-related attacks \cite{Liu2025HarnessingTC}, and ensemble-based attacks \cite{10657707}. Among these approaches, input transformation-based attacks are particularly attractive because of their simplicity, flexibility, and compatibility with standard gradient-based attacks. They construct multiple transformed versions of the current adversarial input and aggregate the gradients calculated from these transformed views, thereby reducing over-reliance on a single input representation of the source surrogate model.

Existing studies mainly explain input transformations from the perspectives of transformed-image diversity, attention-response diversity, or hypothesis-space augmentation \cite{Wang2023BoostingAT,Lin2024BoostingAT}. These explanations provide useful insights, but they primarily focus on image-level diversity or relatively high-level model responses. The interaction between an input transformation and the model frontend, where early feature encoding first occurs, remains insufficiently explored. Visually different transformed inputs do not necessarily induce useful or transferable differences after frontend encoding.

We revisit input transformation-based attacks from an implicit ensemble perspective. A transformation can be regarded as a pre-model operator, and multiple transformed views of the same adversarial input form an implicit ensemble for gradient calculation. Unlike a conventional model ensemble that aggregates gradients from multiple source models, this implicit ensemble applies different pre-model operators to the same source model and aggregates the gradients obtained from the corresponding transformed views.

The effectiveness of such an implicit ensemble depends not only on the visual differences produced by transformations, but also on the frontend responses induced by these differences. If multiple transformed views produce nearly identical frontend responses, their gradients may remain redundant even when the images appear visually different. Conversely, transformations that induce complementary frontend responses can provide more diverse optimization information during gradient aggregation.

In transfer-based attacks, adversarial gradients are calculated only through an accessible source model, while the generated adversarial examples are evaluated on unseen target models. Therefore, transformation-induced differences that are highly specific to the source-model frontend may still lead to source overfitting. In contrast, response variations associated with more general visual structures are more likely to provide useful optimization directions across different architectures. This motivates us to analyze and design input transformations from a frontend response perspective.

Based on this motivation, we propose FRO, a Frontend Response-Oriented input transformation method for transferable adversarial attacks. From an implicit ensemble perspective, different transformed views correspond to composite members $f_s\circ T_i$, and their differences are first encoded by the source-model frontend. FRO aims to perturb frontend factors that are not only diverse in the source model but also partially shared by heterogeneous target architectures.

Specifically, FRO combines a Local Scaling Operator and a Projection Operator. The Local Scaling Operator changes local scale, texture density, boundary positions, and receptive-field or patch-level contents, while the Projection Operator modifies global geometry, object shape, and spatial layout through coherent perspective deformation. At each iteration, the two operators are sequentially sampled to generate $N$ transformed views of $x+\delta_t$, and the corresponding source-model gradients are aggregated to update the adversarial perturbation.

We further define $N$ as the unified implicit ensemble size. Since different transformation-based attacks may use different numbers of transformed views, their performance differences may arise from either transformation design or ensemble scale. Controlling $N$ enables a fairer comparison under the same number of transformed views and gradient evaluations.

Our contributions are summarized as follows.
\begin{itemize}
\item We formulate input transformation-based attacks as a transformed-input implicit ensemble and introduce a frontend response perspective for understanding adversarial transferability.
\item We propose FRO, which combines local scaling and globally coherent projection to induce complementary local and global frontend response variations.
\item We introduce a unified implicit ensemble size $N$ to distinguish the effect of transformation design from that of ensemble scale and to support fairer comparisons among input transformation-based attacks.
\end{itemize}
\section{Related Work}
\subsection{Adversarial Attacks}
\subsubsection{Gradient-based Attacks}
Adversarial attacks generate adversarial examples by applying imperceptible perturbations to input data, causing deep neural networks to make incorrect predictions. Szegedy \etal~\cite{Szegedy2013IntriguingPO} first revealed the vulnerability of DNNs to adversarial examples. Goodfellow \etal~\cite{Goodfellow2014ExplainingAH} proposed the Fast Gradient Sign Method (FGSM), which generates perturbations using a single-step gradient update. Kurakin \etal~\cite{Kurakin2016AdversarialEI} extended FGSM to the iterative I-FGSM, improving white-box attack effectiveness. However, adversarial examples generated in white-box settings often show limited transferability to black-box models~\cite{Liu2016DelvingIT}. Momentum-based attacks, such as MI-FGSM and NI-FGSM, improve transferability by stabilizing the optimization direction~\cite{Dong2017DiscoveringAE,Lin2019NesterovAG}.

Beyond input-space loss optimization, another line of work improves transferability from the feature perspective. ILA~\cite{Huang2019ILA} enhances transferability by amplifying adversarial perturbations along intermediate-layer feature directions. FIA~\cite{Wang2021FIA} exploits object-aware feature importance to reduce model-specific overfitting and improve black-box transferability. These studies suggest that transferable perturbations are related not only to final prediction loss, but also to attack-relevant feature information shared across models.

\subsubsection{Input Transformation-based Attacks}

Input transformation-based attacks have been widely studied to improve adversarial transferability. These methods apply transformations to the current adversarial input and aggregate gradients over transformed views, making the generated perturbation less dependent on a single input pattern or surrogate model. Existing methods design transformations from different perspectives. DIM~\cite{Xie2018ImprovingTO} randomly resizes and pads input images to introduce scale variations, TIM~\cite{Dong2019EvadingDT} exploits translation-invariant gradients, and SIM~\cite{Lin2019NesterovAG} optimizes perturbations over scaled copies of the input. Admix~\cite{Wang2021AdmixET} mixes the input with images from other categories, while CWT~\cite{Liu2025EnhancingAT} and DeCoWA~\cite{Lin2024BoostingAT} introduce constrained variations to preserve semantic or structural consistency. BSR~\cite{Wang2023BoostingAT} shuffles and rotates image blocks to diversify model attention. Beyond individual operators, AITL~\cite{Yuan2021AdaptiveIT}, SIA~\cite{Wang2023StructureIT}, L2T~\cite{Zhu2024LearningTT}, and OPS~\cite{11094976} explore how to combine or select transformation operators during attack optimization. SID~\cite{Zhou2025SID} exploits spatial invariance by constructing multi-scale and multi-position transformed inputs. These studies show that diverse transformed views can improve transferability. However, most existing methods mainly explain transformations from image-level diversity, semantic preservation, attention responses, or operator-combination strategies, while paying less attention to how transformed views are represented by the surrogate-model frontend and how these induced frontend responses participate in gradient aggregation. This motivates us to revisit input transformation-based attacks from an implicit ensemble and frontend-response perspective.

\subsubsection{Interpretations of Input Transformation-based Attacks}

Recent studies have also attempted to explain why input transformation-based attacks improve adversarial transferability. One common explanation is transformed-image diversity, where diverse views reduce dependence on a single input pattern. SIA applies different transformations to segmented image regions, while L2T learns transformation strategies to generate effective transformed views during attack optimization~\cite{Wang2023StructureIT,Zhu2024LearningTT}. Another explanation emphasizes semantic or structural preservation, arguing that useful transformations should not destroy object semantics or global image structure, as reflected in CWT and DeCoWA~\cite{Liu2025EnhancingAT,Lin2024BoostingAT}.

Another line of interpretation focuses on model-side responses induced by transformed inputs. BSR explains transferability through attention-response diversity, where block shuffling and rotation diversify model attention patterns~\cite{Wang2023BoostingAT}. OPS treats input transformations as pre-processing operators inserted before the surrogate model, forming diverse composite mappings and augmenting the implicit hypothesis space~\cite{11094976}. These studies suggest that input transformations are not only image-level operations, but also affect how the surrogate model responds to transformed views.

Recent analyses further show that input transformations can influence internal feature responses. Input Transpose observes that simple input-level transformations can induce low-level feature-map fluctuations~\cite{InputTranspose}. FAUG studies transferability from the feature-augmentation perspective by perturbing intermediate representations~\cite{Wang2025FAUG}. Frequency- and multi-scale-based methods, such as FSA and DEM, further suggest that transformed inputs or transformed gradients may affect model responses across different frequency bands or spatial scales~\cite{Zheng2025FSA,Zou2020RDIM}. Overall, existing interpretations cover image diversity, semantic or structural preservation, attention-response diversity, hypothesis-space augmentation, internal feature variations, and frequency/multi-scale responses.

\subsection{Adversarial Defense}
To mitigate adversarial attacks, researchers have developed various defense strategies. Madry \etal, proposed PGD-based Adversarial Training (AT), which improves robustness but incurs high computational costs \cite{Madry2017TowardsDL}. Tramér \etal, introduced Ensemble Adversarial Training \cite{tramer2018ensemble}, resulting in robust models such as ens3\_adv\_Inception-V3, ens4\_adv\_Inception-V3, and ens2\_adv\_Inception-ResNet. Xu \etal, proposed feature squeezing, including bit-depth reduction (Bit-RD), to compress input features and detect adversarial examples \cite{Xu2017FeatureSD}. Liao \etal, introduced high-level representation-guided denoisers (HGD) to purify adversarial perturbations \cite{Liao2017DefenseAA}. Cohen \etal, employed randomized smoothing (RS) to train certifiably robust classifiers \cite{Cohen2019CertifiedAR}. Wong \etal, developed a single-step adversarial training method to reduce computational cost while maintaining robustness \cite{Wong2020FastIB}. Naseer \etal, proposed Neural Representation Purifiers (NRP) against transferable adversarial attacks \cite{Naseer2020ASA}. Nie \etal, proposed DiffPure, which uses diffusion models to remove adversarial perturbations through forward diffusion and reverse denoising \cite{Nie2022DiffPure}.

\section{Methods}

\subsection{Preliminary Study: Transformation-induced Front-end Responses}
\label{sec:preliminary_study}

Input transformation-based attacks have been widely used to improve adversarial transferability. Existing explanations involve attention-response diversity, semantic preservation, hypothesis-space augmentation, transformation-distribution exploration, and cross-model feature consistency~\cite{Xie2018ImprovingTO,Dong2019EvadingDT,Wang2023StructureIT,Wang2023BoostingAT,Lin2024BoostingAT,11094976,Hu2025TVIA,Liu2025EnhancingAT,Chen2025SAA}. These studies suggest that the effect of input transformations cannot be fully characterized by input appearance alone. Motivated by this, we examine how transformed views are encoded by heterogeneous front-ends.

Specifically, for each clean image \(x\), we generate transformed views \(\{T_i(x)\}_{i=1}^{N}\), and feed both the clean image and transformed views into a CNN model and a ViT model. We then extract their early front-end features and convert them into response maps. For CNN features with shape \([C,H,W]\), we average the channel-wise activation magnitude to obtain a spatial response map. For ViT features with shape \([P,C]\), where \(P\) denotes the number of patch tokens, we remove the class token, reshape patch tokens into a spatial grid, and average the embedding-wise activation magnitude. The resulting response maps are upsampled and overlaid on the corresponding inputs for qualitative comparison.

\begin{figure}[htbp]
    \centering
    \includegraphics[width=0.48\textwidth]{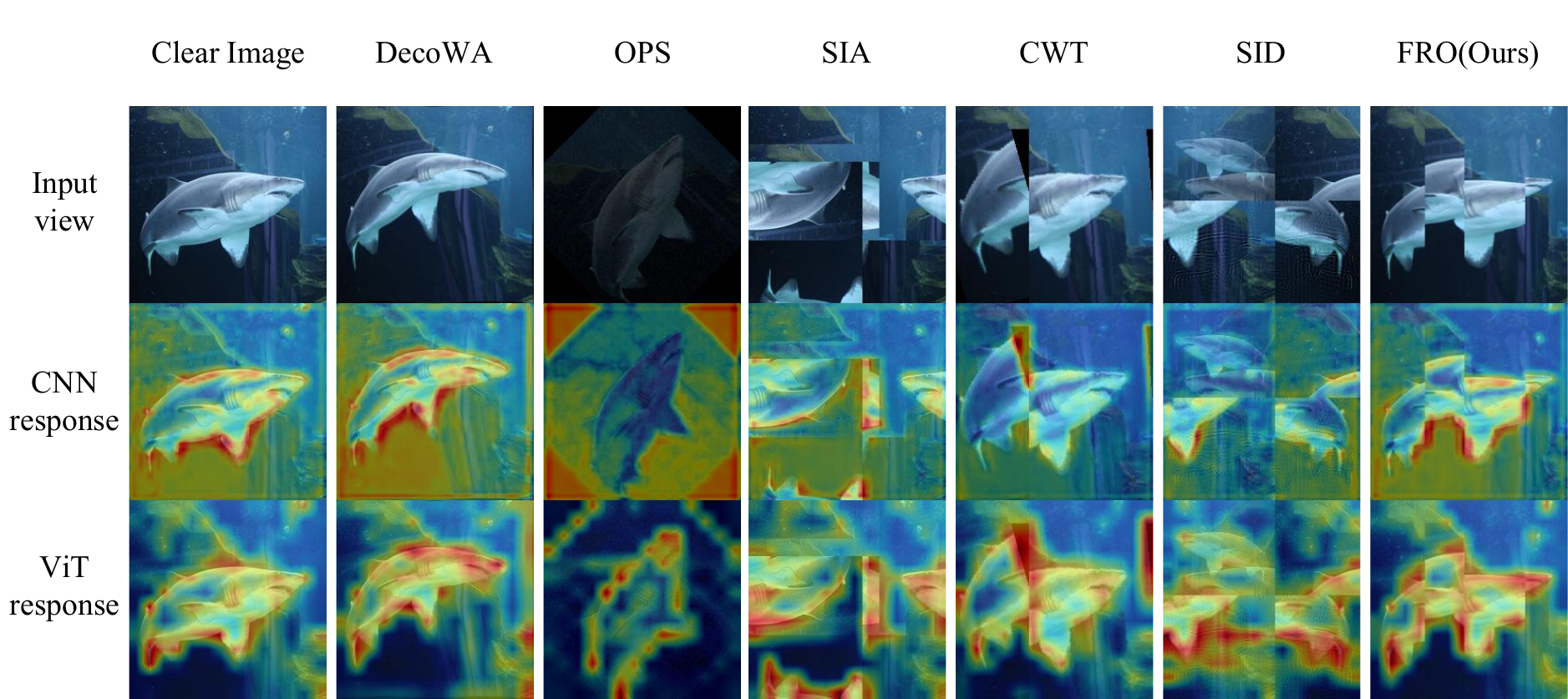}
    \caption{Preliminary study of transformation-induced front-end responses. For the same clean image and transformed views, CNN and ViT front-ends often emphasize partially overlapping object-related regions. Transformation-induced CNN response changes can also be partially reflected in the ViT front-end around similar object parts, boundaries, or local structures.}
    \label{fig:frontend_response_observation}
\end{figure}

As shown in Fig.~\ref{fig:frontend_response_observation}, CNN and ViT front-ends often emphasize partially overlapping object-related regions, such as salient object parts, boundaries, high-contrast textures, and local structures. More importantly, this partial correspondence remains visible under transformed views. Different transformed views induce different response regions in the CNN front-end, and some of these changes are also reflected in the ViT front-end around similar object parts or structural cues.

These observations indicate that transformation-induced source-front-end response differences may contain spatial or structural factors that are also reflected by a heterogeneous target front-end. Therefore, we further examine whether the response relations induced among transformed views are preserved across source and target front-ends.

\subsection{Motivation}
\label{sec:motivation}

The preliminary study suggests that transformed views may induce source-front-end response differences that are not completely isolated from heterogeneous target front-ends. We next explain why this observation is relevant to adversarial transferability.

Our motivation starts from the ensemble view of transfer attacks. Ensemble-based attacks improve transferability by aggregating gradients from multiple members, where member diversity provides complementary optimization directions and aggregation tends to retain attack-relevant components that are stable across members while suppressing member-specific components~\cite{Liu2016DelvingIT,Dong2017DiscoveringAE}. Thus, transferability is not determined by diversity alone, but by whether the optimized perturbation preserves components that remain effective beyond a particular source member.

Input transformation-based attacks can be interpreted as a functional implicit ensemble. Given a source model \(f_s\) and transformations \(\{T_i\}_{i=1}^{N}\), the attack optimizes
\begin{equation}
\max_{\delta}
\frac{1}{N}
\sum_{i=1}^{N}
\mathcal{L}
\left(
f_s(T_i(x+\delta)), y
\right),
\end{equation}
which corresponds to a set of transformed source members:
\begin{equation}
\mathcal{E}_{s}
=
\{f_s\circ T_i\}_{i=1}^{N}.
\end{equation}
Here, input transformations do not create independently trained models. Each implicit member is the composite function \(f_s\circ T_i\).

To analyze how transformation-induced differences are perceived by the source model, we decompose the source model as
\begin{equation}
f_s = g_s\circ h_s,
\end{equation}
where \(h_s\) is the source-model front-end feature extractor and \(g_s\) is the remaining classifier. Each transformed member becomes
\begin{equation}
f_s\circ T_i
=
g_s\circ h_s\circ T_i .
\end{equation}
Thus, \(T_i\) introduces input variation, while \(h_s\circ T_i\) describes how this variation is encoded into source-front-end responses.

Inspired by the ensemble view above, we further clarify where member differences arise in input transformation-based attacks. In conventional ensemble attacks, different members usually correspond to different model parameters or architectures. Their diversity provides complementary gradient directions, and gradient aggregation can reduce overfitting to a single source member, thereby improving adversarial transferability. In contrast, input transformation-based attacks use the same source model \(f_s\) for all transformed views. Therefore, the implicit members \(\{f_s\circ T_i\}_{i=1}^{N}\) do not differ in independently learned model parameters.

After decomposing \(f_s=g_s\circ h_s\), each implicit member can be written as \(g_s\circ h_s\circ T_i\). Since \(g_s\) is shared across all implicit members, the effective member differences mainly arise from the transformation \(T_i\) and how the shared source front-end \(h_s\) encodes the transformed input. In this sense, the diversity of the implicit ensemble is not model-parameter diversity, but transformation-induced front-end response diversity.

Based on this view, we do not claim that front-end responses themselves are the direct source of transferability. Instead, we argue that transformation-induced front-end response differences constitute the effective differences among implicit ensemble members. When these differences provide non-redundant gradient directions and are not purely source-specific, their aggregation is more likely to produce perturbation components that remain effective across architectures. We refer to the response factors associated with such cross-architecture relevance as \emph{target-relevant response factors}.

In the following analysis, we instantiate this formulation in a CNN-to-ViT transfer setting, where \(f_s=f_{\mathrm{CNN}}\) and \(f_t=f_{\mathrm{ViT}}\). The target ViT is used only for diagnostic analysis and is never involved in perturbation optimization. Since CNN and ViT activations have different forms and dimensions, we do not directly align raw features. Instead, we compare the response relations induced among the same transformed views within each front-end. If the response relation formed by
\begin{equation}
\{h_{\mathrm{CNN}}(T_i(x))\}_{i=1}^{N}
\end{equation}
has a smaller gap to the response relation formed by
\begin{equation}
\{h_{\mathrm{ViT}}(T_i(x))\}_{i=1}^{N},
\end{equation}
then the transformation-induced source-front-end differences are more likely to contain cross-front-end-preserved response factors.

It is worth noting that cross-front-end response preservation is not a sufficient condition for transferability. A small response-geometry gap is meaningful only when the transformed views also induce non-trivial source-side response diversity. Therefore, the motivation above does not directly prescribe a specific transformation operator. Instead, it points to a design principle: input transformations should perturb front-end response factors that are likely to be meaningful for both the source and target architectures. In the next subsection, we identify such shared front-end factors and explain how they motivate the design of FRO.

\subsection{Shared Front-end Factors for Cross-architecture Consistency}
\label{sec:shared_frontend_factors}

To make the above motivation operational, we identify two front-end factors that are commonly involved in both CNN and ViT feature encoding: local content sampling and spatial organization. These factors are not intended to claim that CNNs and ViTs process images in the same way. Instead, they describe two common entry points through which an input transformation can affect the early encoding behavior of both architectures. This view is consistent with representation studies showing that ViTs differ from CNNs in their internal representation structure, while still preserving spatial information~\cite{Raghu2021DoVT}.

The first factor is local content sampling. A CNN front-end aggregates local patterns through convolutional receptive fields, so its response at each spatial location is affected by local scale, texture density, boundary position, and the arrangement of nearby structures. A ViT front-end follows a different mechanism, but it also starts from local patch tokens by partitioning the image into patches and embedding them into a token sequence~\cite{Dosovitskiy2020ViT}. Therefore, the content contained within each patch region, as well as the distribution of local structures across neighboring patches, can also affect the early ViT response. Although CNNs and ViTs organize features differently, both front-ends are sensitive to how local contents are sampled and presented.

The second factor is spatial organization. CNNs preserve spatial structure through feature maps, where neighboring activations correspond to neighboring image regions. ViTs preserve spatial information through patch order and positional encoding, and self-attention further models relationships among patch tokens. Window-based and hierarchical vision Transformers also emphasize local windows, shifted-window interactions, and multi-scale spatial organization~\cite{Liu2021Swin,Dong2022CSWin}. Therefore, global geometry, object shape, part arrangement, and low-frequency layout can influence the front-end responses of both architectures. Transformations that modify spatial organization in a globally coherent manner may induce response changes that are less tied to one specific architecture.

Based on these considerations, FRO designs two complementary operators from the above shared front-end factors. The Local Scaling Operator targets local content sampling by stretching and shrinking local image regions, thereby changing local scale, texture density, boundary position, and receptive-field or patch-level contents while preserving the overall image size. The Projection Operator targets spatial organization by applying a globally coherent perspective deformation, thereby changing global geometry, object shape, and spatial layout. These two operators are not used to directly optimize the target model or minimize the cross-front-end gap during perturbation generation. Rather, they aim to induce non-redundant source-front-end response variations that are more likely to contain cross-front-end-consistent response factors.

\subsection{Frontend Response-oriented Transformation Operators}
\label{sec:frontend_response_transformation_operators}

\begin{figure*}[htbp]
    \centering
    \includegraphics[width=1\textwidth]{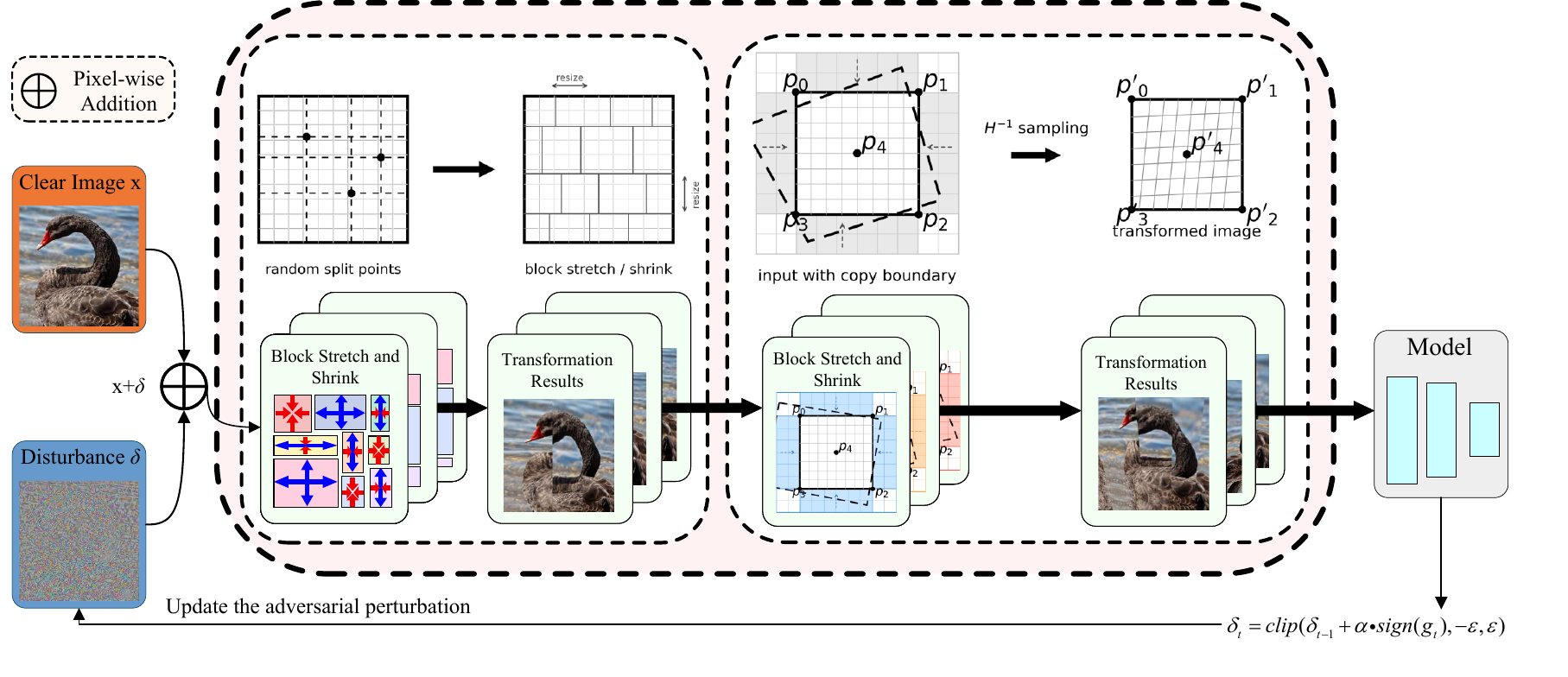}
    \caption{Overview of the frontend-response-oriented transformed-view generation framework. Given a clean image \(x\) and adversarial perturbation \(\delta\), the perturbed input \(x+\delta\) is processed by complementary transformation operators, including local block stretch-and-shrink and globally coherent projection deformation. The resulting transformed views are fed into the source model to obtain gradient components, which are aggregated to update the adversarial perturbation.}
    \label{fig:1vsob41}
\end{figure*}

Based on the shared front-end factors discussed above, we construct transformed views with two complementary operators: a Local Scaling Operator and a Projection Operator. The Local Scaling Operator targets local content sampling by changing local scale, texture density, boundary positions, and receptive-field or patch-level contents. The Projection Operator targets spatial organization by introducing globally coherent geometric deformation. Together, these operators are expected to induce non-redundant front-end response variations for gradient aggregation.

\subsubsection{Local Scaling Operator}
\label{sec:local_scaling_operator}

To induce local front-end response variations, we introduce a Local Scaling Operator that stretches and shrinks image regions along the spatial dimensions. Instead of applying a global transformation to the whole image, this operator first partitions the input into several local blocks and then randomly adjusts their spatial lengths while preserving the overall image size. In this way, local scale, texture density, boundary position, and receptive-field or patch-level contents are changed without destroying the global image layout.

Given an adversarial input \(x+\delta \in \mathbb{R}^{C\times H\times W}\), we first sample \(M\) segmentation points with two constraints: each point should keep at least a border distance \(d_b\) from the image boundary, and any two segmentation points should be separated by at least \(d_p\). For a spatial dimension with length \(L\), the sorted segmentation points are denoted as
\begin{equation}
0 = p^{\mathrm{sort}}_0 
< p^{\mathrm{sort}}_1 
< \cdots 
< p^{\mathrm{sort}}_M 
< p^{\mathrm{sort}}_{M+1}=L .
\end{equation}
These points divide the input into \(M+1\) blocks:
\begin{equation}
B_i =
\operatorname{slice}
\left(
x+\delta,\,
p^{\mathrm{sort}}_{i-1}:p^{\mathrm{sort}}_{i}
\right),
\quad i=1,\dots,M+1,
\end{equation}
with the original block length
\begin{equation}
l_i = p^{\mathrm{sort}}_{i}-p^{\mathrm{sort}}_{i-1}.
\end{equation}
Here, \(M\) is the number of segmentation points, \(d_b\) is the border distance, \(d_p\) is the minimum interval between segmentation points, and \(L\) denotes the length of the current spatial dimension.

For each block, we sample a random scaling score \(s_i\) from a uniform distribution controlled by the range-to-center ratio \(r\):
\begin{equation}
s_i \sim 
\mathcal{U}
\left(
\frac{1-r/2}{2},
\frac{1+r/2}{2}
\right),
\quad i=1,\dots,M+1 .
\end{equation}
The scores are normalized to obtain the target length proportion:
\begin{equation}
w_i = \frac{s_i}{\sum_{j=1}^{M+1}s_j}.
\end{equation}
The target length of each block is then computed as
\begin{equation}
\hat{l}_i = \operatorname{round}(L\cdot w_i).
\end{equation}
An adjustment operation \(\operatorname{Adj}(\cdot)\) ensures that all target lengths are positive integers and their sum equals \(L\):
\begin{equation}
(l_1',\dots,l_{M+1}')
=
\operatorname{Adj}(\hat{l}_1,\dots,\hat{l}_{M+1}),
\quad
\sum_{i=1}^{M+1} l_i'=L .
\end{equation}
Here, \(r\) controls the sampling range of scaling scores, \(w_i\) is the normalized target length proportion, \(\hat{l}_i\) is the rounded target length, and \(l_i'\) is the adjusted target length.

Each block is resized to its adjusted target length by bilinear interpolation:
\begin{equation}
B_i' = \operatorname{Interp}(B_i, l_i'),
\end{equation}
and the transformed output along this dimension is obtained by concatenating all resized blocks:
\begin{equation}
\tilde{x}
=
\operatorname{Concat}
(B_1',B_2',\dots,B_{M+1}').
\end{equation}
The same operation can be applied along both height and width dimensions to generate the final locally scaled transformed image.

This operator preserves the overall image size while introducing local stretch-and-shrink variations. As a result, it provides transformed views that induce front-end response variations associated with local scale, texture density, and receptive-field or patch-level structures.

\subsubsection{Projection Operator}
\label{sec:projection_operator}

To induce global front-end response variations, we introduce a Projection Operator that applies a coherent perspective deformation to the entire image. Given an adversarial input \(x+\delta \in \mathbb{R}^{C\times H\times W}\), we represent the image plane using four normalized corner points
\begin{equation}
\mathcal{P}=\{(-1,-1),(1,-1),(1,1),(-1,1)\}.
\end{equation}

A projection instance is controlled by a tilt angle \(\alpha\), a projection-axis angle \(\theta\), and a virtual focal length \(f_p\), where
\begin{equation}
\alpha \sim \mathcal{U}(-\alpha_{\max},\alpha_{\max}),\quad
\theta \sim \mathcal{U}(0,2\pi),\quad
\alpha_{\max}=\frac{\pi}{180}A_{\max}.
\end{equation}
Here, \(A_{\max}\) controls the maximum tilt angle and \(f_p\) controls the strength of the perspective effect. These parameters define a perspective projection function \(\Pi_{\alpha,\theta,f_p}(\cdot)\), which maps the source corner set \(\mathcal{P}\) to a target corner set
\begin{equation}
\mathcal{P}'=\{\mathbf{p}'_i\}_{i=1}^{4},\quad
\mathbf{p}'_i=\Pi_{\alpha,\theta,f_p}(\mathbf{p}_i),\quad
\mathbf{p}_i\in\mathcal{P}.
\end{equation}

Unlike independently perturbing the four corners, this projection is generated from a shared geometric parameterization, so the resulting deformation remains globally coherent.

Given the corner correspondence \(\mathcal{P}\rightarrow\mathcal{P}'\), we estimate a homography matrix \(\mathbf{H}\in\mathbb{R}^{3\times3}\) by Direct Linear Transform. The homography satisfies
\begin{equation}
\lambda_i
\begin{bmatrix}
\mathbf{p}'_i \\
1
\end{bmatrix}
=
\mathbf{H}
\begin{bmatrix}
\mathbf{p}_i \\
1
\end{bmatrix},
\quad i=1,\dots,4,
\end{equation}
where \(\lambda_i\) is a non-zero scale factor.

The transformed image is then generated by inverse warping. For each output coordinate \(\mathbf{q}'=(u',v',1)^\top\), the corresponding source coordinate is computed as
\begin{equation}
\begin{aligned}
\mathbf{q} &= \mathbf{H}^{-1}\mathbf{q}' \\
(u,v) &=
\left(
\frac{q_x}{q_z},
\frac{q_y}{q_z}
\right).
\end{aligned}
\end{equation}

The output pixel is obtained by bilinear sampling from \(x+\delta\). During inverse warping, some source coordinates may fall outside the valid image domain. We handle these locations with reflection padding before bilinear sampling:
\begin{equation}
(u_r,v_r)=\operatorname{Reflect}(u,v),
\quad
\tilde{x}(u',v')=
\operatorname{Bilinear}
\left(
x+\delta,
(u_r,v_r)
\right).
\end{equation}
For coordinates inside the valid domain, \((u_r,v_r)=(u,v)\). This boundary handling avoids black artifacts caused by zero padding and keeps the image boundary visually continuous.

Since the same homography is applied to all pixels, the generated view changes the global geometry, spatial layout, object shape, and low-frequency structure of the input. As a result, it provides transformed views that induce front-end response variations associated with global geometry and spatial layout, complementing the local response variations introduced by the Local Scaling Operator.

\subsubsection{Transformed-view Generation}
\label{sec:transformed_view_generation}

At the \(t\)-th iteration, we apply the Local Scaling Operator and the Projection Operator to the current perturbed image \(x+\delta_t\). After sampling these two operators \(N\) times, we obtain the transformed-view set:
\begin{equation}
\mathcal{T}_N(x+\delta_t)
=
\left\{
\tilde{x}^{(k)}_t
\right\}_{k=1}^{N}.
\end{equation}

Each transformed sample is generated by sequentially applying the Local Scaling Operator and the Projection Operator:
\begin{equation}
\tilde{x}^{(k)}_t
=
T_{\mathrm{Proj}}^{(k)}
\left(
T_{\mathrm{LS}}^{(k)}(x+\delta_t)
\right),
\quad k=1,\dots,N.
\end{equation}
Here, \(T_{\mathrm{LS}}^{(k)}\) denotes the \(k\)-th sampled Local Scaling Operator, and \(T_{\mathrm{Proj}}^{(k)}\) denotes the \(k\)-th sampled Projection Operator. The resulting set \(\mathcal{T}_N(x+\delta_t)\) is used for gradient aggregation in the current iteration.

\subsection{Perturbation Generation}
\label{sec:perturbation_generation}

Given the transformed-view set \(\mathcal{T}_N(x+\delta_t)=\{\tilde{x}^{(k)}_t\}_{k=1}^{N}\), we compute the average loss over all transformed samples:
\begin{equation}
\mathcal{J}_t
=
\frac{1}{N}
\sum_{k=1}^{N}
J
\left(
f_s(\tilde{x}^{(k)}_t), y
\right).
\end{equation}
The corresponding average gradient is
\begin{equation}
\bar{\nabla}_t
=
\nabla_{\delta}\mathcal{J}_t
=
\frac{1}{N}
\sum_{k=1}^{N}
\nabla_{\delta}
J
\left(
f_s(\tilde{x}^{(k)}_t), y
\right),
\end{equation}
where \(f_s\) is the source model, \(J(\cdot,\cdot)\) is the attack loss, \(y\) is the ground-truth label, and \(N\) is the number of transformed views used for gradient aggregation.

We update the perturbation using MI-FGSM. The perturbation and accumulated momentum are initialized as \(\delta_0=0\) and \(g_0=0\). At the \(t\)-th iteration, the momentum is updated by
\begin{equation}
g_t
=
\mu g_{t-1}
+
\frac{\bar{\nabla}_t}
{\|\bar{\nabla}_t\|_1},
\end{equation}
where \(\mu\) is the momentum decay factor. The perturbation is then updated as
\begin{equation}
\delta_{t+1}
=
\operatorname{clip}_{[-\epsilon,\epsilon]}
\left(
\delta_t
+
\eta \cdot \operatorname{sign}(g_t)
\right),
\end{equation}
where \(\epsilon\) is the perturbation budget and \(\eta=\epsilon/T\) is the step size. After \(T\) iterations, the final adversarial example is
\begin{equation}
x_{\mathrm{adv}}
=
\operatorname{clip}_{[0,1]}
\left(
x+\delta_T
\right).
\end{equation}

\begin{figure*}[t]
    \centering
    \includegraphics[width=0.95\textwidth]{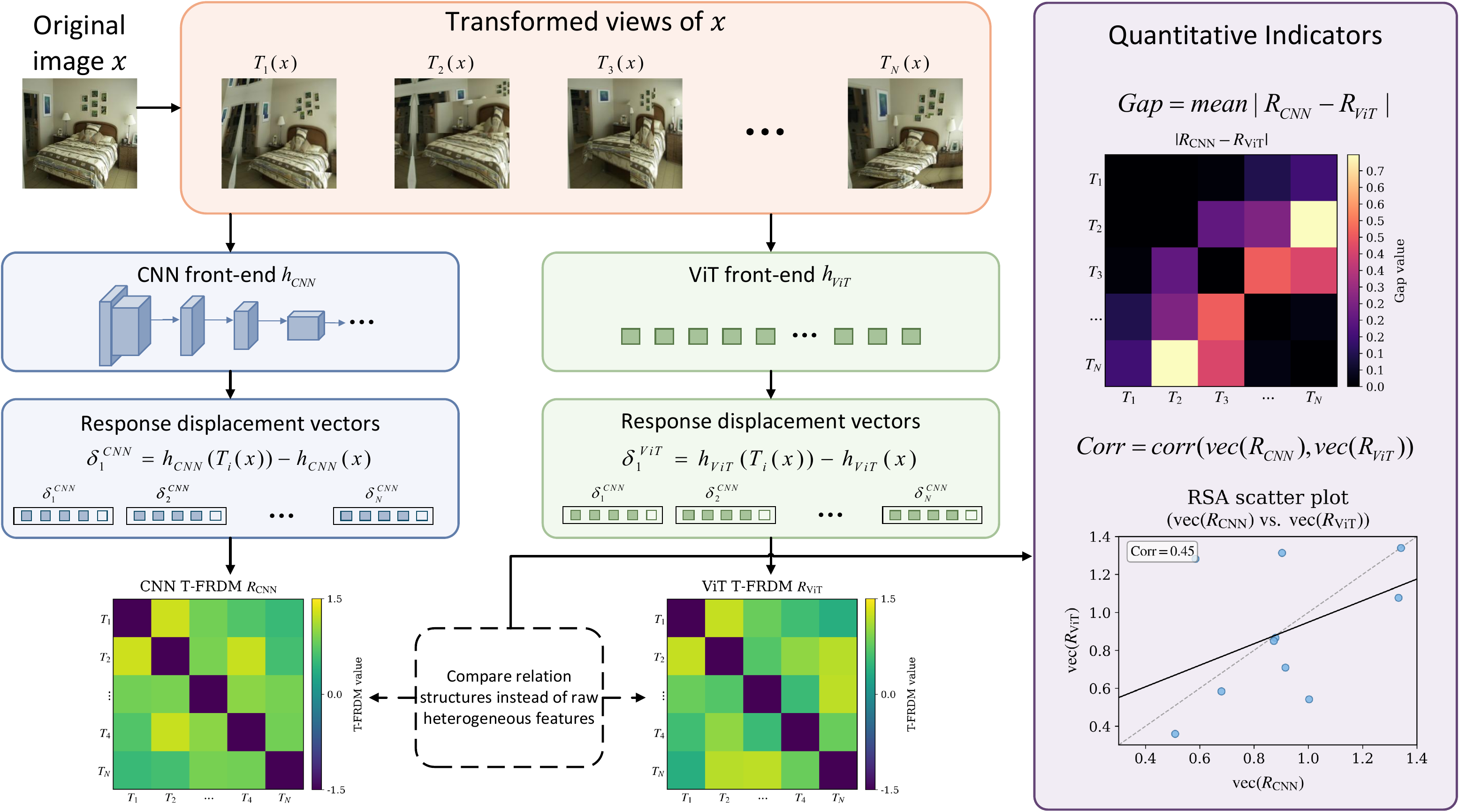}
    \caption{
    Illustration of the RSA-based CNN--ViT front-end response relation gap.
    The same transformed views are fed into representative CNN-based and ViT-based front-ends.
    Instead of comparing raw heterogeneous features, we construct transformation-induced front-end response dissimilarity matrices and compare their relation structures using coupled gap and correlation.
    }
    \label{fig:gapfunction}
\end{figure*}

\subsection{Transformation-induced Frontend Response Relation Gap}
\label{sec:frontend_gap}

Input transformation-based attacks usually generate multiple transformed views to improve transferability. However, the effect of these transformed views is closely related to how they interact with the front-end representations of different architectures. In this work, we focus on the response discrepancy between CNN-based and ViT-based front-ends. Since these two types of architectures organize visual information in different forms, directly comparing their raw feature activations is not suitable. CNN-based front-ends usually produce convolutional feature maps with local spatial aggregation, while ViT-based front-ends represent images through patch tokens and Transformer blocks. Therefore, their feature dimensions, spatial organizations, and response scales are not naturally aligned.

To compare heterogeneous front-ends in a unified way, we introduce a relation-level comparison based on representational similarity analysis. Instead of directly aligning raw front-end features, we compare the response relation structures induced by the same set of transformed views. Let \(x\) denote the original image, and let \(\{T_i(x)\}_{i=1}^{N}\) denote a set of transformed views generated from \(x\). We use \(h_{\mathrm{CNN}}\) and \(h_{\mathrm{ViT}}\) to denote representative CNN-based and ViT-based front-end operators, respectively. In our implementation, they are instantiated by the front-end layer of ResNet-18 and the early block of ViT-B/16.

For each transformed view \(T_i(x)\), we first compute its front-end response displacement with respect to the original image:
\begin{equation}
\Delta_i^{h}
=
h(T_i(x))-h(x),
\end{equation}
where \(h \in \{h_{\mathrm{CNN}}, h_{\mathrm{ViT}}\}\). The displacement \(\Delta_i^{h}\) describes how the \(i\)-th input transformation changes the front-end response of the original image under a given architecture.

Based on these response displacements, we construct a transformation-induced front-end response dissimilarity matrix, denoted as T-FRDM. For a given front-end \(h\), each element of the matrix is defined as
\begin{equation}
R_h[i,j]
=
d(\Delta_i^{h},\Delta_j^{h}),
\end{equation}
where \(d(\cdot,\cdot)\) measures the response-relation distance between the \(i\)-th and \(j\)-th transformed views.

Considering that CNN-based and ViT-based front-ends may have different response scales, we adopt a normalized coupled response distance:
\begin{equation}
d(\Delta_i^{h},\Delta_j^{h})
=
\frac{
\operatorname{mean}_{s,c}
\left(
\Delta_i^{h}(s,c)-\Delta_j^{h}(s,c)
\right)^2
}{
\tau_h+\epsilon_{\mathrm{num}}
},
\end{equation}
where \(s\) denotes the spatial location for convolutional feature maps or the patch-token index for ViT features, \(c\) denotes the channel or embedding dimension, and \(\epsilon_{\mathrm{num}}\) is a small constant for numerical stability. The normalization factor \(\tau_h\) is computed as
\begin{equation}
\tau_h
=
\operatorname{mean}_{i,s,c}
\left(
\Delta_i^{h}(s,c)
\right)^2.
\end{equation}
This normalization reduces the influence of architecture-dependent response-scale differences. Since the distance term jointly reflects response magnitude and response direction, the coupled response distance provides an integrated description of the front-end response relation induced by input transformations.

For the same set of transformed views, we obtain two T-FRDMs:
\begin{equation}
R_{\mathrm{CNN}}
=
R_{h_{\mathrm{CNN}}},
\quad
R_{\mathrm{ViT}}
=
R_{h_{\mathrm{ViT}}}.
\end{equation}
Although the original feature spaces of CNN-based and ViT-based front-ends are heterogeneous, \(R_{\mathrm{CNN}}\) and \(R_{\mathrm{ViT}}\) have the same size \(N \times N\), because they both describe pairwise relations among the same transformed views. Therefore, the comparison is performed in the relation space rather than the raw feature space.

We then define the coupled front-end response relation gap as
\begin{equation}
\mathrm{Gap}
=
\operatorname{mean}_{i\neq j}
\left|
R_{\mathrm{CNN}}[i,j]
-
R_{\mathrm{ViT}}[i,j]
\right|.
\end{equation}
A smaller \(\mathrm{Gap}\) indicates that the response relations induced in the CNN-based front-end are more consistently preserved in the ViT-based front-end.

In addition, we compute the response-relation correlation:
\begin{equation}
\mathrm{Corr}
=
\operatorname{corr}
\left(
\{R_{\mathrm{CNN}}[i,j]\}_{i\neq j},
\{R_{\mathrm{ViT}}[i,j]\}_{i\neq j}
\right).
\end{equation}
A higher \(\mathrm{Corr}\) indicates that the two front-ends preserve more similar relative relation structures among the transformed views.

Figure~\ref{fig:gapfunction} illustrates the proposed RSA-based front-end response relation gap. Given the same transformed views, we compute their response displacements in CNN-based and ViT-based front-ends, construct the corresponding T-FRDMs, and then measure the discrepancy and correlation between the two relation matrices. This design avoids directly comparing heterogeneous raw features and instead compares transformation-induced response relation structures. Since the coupled response distance jointly captures response direction and response magnitude, it provides an integrated measure of cross-architecture front-end response consistency.

\subsection{Evaluating Input Transformation-based Methods under a Unified Implicit Ensemble Size}
\label{sec:Unified Num Scale}

Following the front-end-response ensemble formulation in Sec.~\ref{sec:motivation}, we define the \emph{implicit ensemble size} \(N\) as the number of transformed views used for gradient aggregation at each optimization step. Given the current perturbed input \(x+\delta_t\), an attack constructs
\begin{equation}
\mathcal{T}_N(x+\delta_t)
=
\left\{
T_i(x+\delta_t)
\right\}_{i=1}^{N},
\end{equation}
which induces the front-end response set
\begin{equation}
\mathcal{H}_N(x+\delta_t)
=
\left\{
h(T_i(x+\delta_t))
\right\}_{i=1}^{N}.
\end{equation}
The averaged gradient is computed as
\begin{equation}
\bar{g}_t
=
\frac{1}{N}
\sum_{i=1}^{N}
\nabla_{\delta}
\mathcal{L}
\left(
g(h(T_i(x+\delta_t))), y
\right).
\end{equation}
Thus, \(N\) controls both the number of transformed views and the number of front-end responses involved in each update.

Using a unified \(N\) is necessary for fair comparison. Otherwise, performance differences may come from ensemble scale rather than transformation design. A larger \(N\) introduces more transformed views, front-end responses, and model evaluations, making it difficult to attribute gains to the transformation strategy itself.

The value of \(N\) also reflects how a transformation family scales. Complementary views may improve gradient stability, redundant views may lead to saturation, and excessively distorted or weakly related views may weaken the averaged gradient. Therefore, unless otherwise specified, all input transformation-based methods are evaluated under the same implicit ensemble size \(N\), with runtime and memory cost reported separately.
\section{Experiments}

\subsection{Experimental Setup}
\label{sec:experimental_setup}

\textbf{Hardware and attack settings}: All experiments are conducted on an NVIDIA RTX 3090 GPU with 24 GB VRAM. Unless otherwise specified, we use random seed \(42\), attack iterations \(T=10\), batch size \(B=1\), perturbation budget \(\|\delta\|_{\infty}\le \epsilon\) with \(\epsilon=16/255\), and step size \(\eta=\epsilon/T=1.6/255\). MI-FGSM with momentum decay factor \(\mu=1\) is used as the base optimizer. All attacks are untargeted. Adversarial examples are generated on a white-box surrogate model and evaluated on held-out black-box target models.

\textbf{Models}: We use ResNet-18~\cite{He2015DeepRL}, ResNet-101~\cite{He2015DeepRL}, DenseNet-121~\cite{Huang2016DenselyCC}, Inception-v3~\cite{Szegedy2015RethinkingTI}, and ViT-B/16~\cite{Dosovitskiy2020AnII} as source models. Target models include six CNNs, namely ResNet-18, ResNet-50, ResNeXt-50, DenseNet-121, Inception-v3, and Inception-v4~\cite{Szegedy2016Inceptionv4IA}, and five Transformer-based models, namely ViT-B/16, PiT~\cite{Heo2021RethinkingSD}, Visformer~\cite{Chen2021VisformerTV}, Swin~\cite{Liu2021SwinTH}, and CaiT~\cite{Touvron2021GoingDW}. When the source model also appears in the target-model set, its white-box result is excluded from black-box averages.

\textbf{Baselines}: We compare FRO with representative input transformation-based attacks, including DeCoWA, CWT, SIA, SID, and OPS. All methods are integrated into MI-FGSM with identical attack hyperparameters. CWT is evaluated using the official implementation, while DeCoWA, SIA, SID, and OPS are adopted from the \texttt{TransferAttack} library without modification. Additional implementation details are provided in Appendix~\ref{app:implementation_details}.

\textbf{Defended models}: Defense experiments are conducted using Inception-v3 as the white-box source model. The generated adversarial examples are evaluated on three ensemble adversarially trained models, namely ens3\_adv\_Inception-v3, ens4\_adv\_Inception-v3, and ens2\_adv\_Inception-ResNet~\cite{tramer2018ensemble}, as well as representative defense methods including AT~\cite{Madry2017TowardsDL}, HGD~\cite{Liao2017DefenseAA}, NRP~\cite{Naseer2020ASA}, and DiffPure~\cite{Nie2022DiffPure}.

\subsection{Effect of the Number of Transformed Views}
\label{sec:number_scale_analysis}

We study how the number of transformed views \(N\) affects the performance of input transformation-based attacks. In our frontend-response ensemble formulation, \(N\) controls both the number of transformed inputs used for gradient aggregation and the number of induced frontend-response members. Therefore, varying \(N\) directly changes the implicit ensemble scale of an attack.

Following the unified implicit ensemble size defined in Sec.~\ref{sec:Unified Num Scale}, we evaluate \(N \in \{10,17,21,26,37,101\}\). To avoid using a large table in the main text, we visualize the attack success rates under different number scales in \cref{fig:numscale}. The complete numerical results are provided in Appendix~\ref{app:number_scale_full}.


\definecolor{siacolor}{HTML}{EE7733}
\definecolor{decowacolor}{HTML}{EE3377}
\definecolor{opscolor}{HTML}{228833}
\definecolor{cwtcolor}{HTML}{4477AA}
\definecolor{sidcolor}{HTML}{AA4499}
\definecolor{frocolor}{HTML}{CC3311}

\pgfplotsset{compat=1.18}
\pgfplotsset{
  every axis/.append style={
    height=4.35cm,
    width=4.35cm,
    xlabel style={font=\scriptsize},
    ylabel style={font=\scriptsize, inner sep=1pt},
    tick label style={font=\scriptsize},
    grid style={dashed,gray!60},
    axis line style={draw=none},
    tick align=outside,
    clip=false,
    ylabel shift={-2pt},
    xticklabel style={rotate=45, anchor=east, font=\scriptsize},
    yticklabel style={font=\scriptsize},
  },
  sia/.style={mark=square*, mark size=1.4pt, color=siacolor, line width=0.9pt, smooth},
  decowa/.style={mark=triangle*, mark size=1.6pt, color=decowacolor, line width=0.9pt, smooth},
  ops/.style={mark=diamond*, mark size=1.5pt, color=opscolor, line width=0.9pt, smooth},
  cwt/.style={mark=*, mark size=1.4pt, color=cwtcolor, line width=0.9pt, smooth},
  sid/.style={mark=star, mark size=1.9pt, color=sidcolor, line width=0.9pt, smooth},
  fro/.style={mark=pentagon*, mark size=1.6pt, color=frocolor, line width=1.0pt, smooth},
}

\begin{figure*}[!t]
  \centering

  \begin{tikzpicture}
    \begin{axis}[
      hide axis,
      xmin=0, xmax=1, ymin=0, ymax=1,
      legend style={
        at={(0.5,1.0)},
        anchor=north,
        draw=none,
        fill=none,
        font=\scriptsize,
        legend columns=6,
        column sep=8pt,
      },
    ]
      \addlegendimage{sia} \addlegendentry{SIA}
      \addlegendimage{decowa} \addlegendentry{DeCoWA}
      \addlegendimage{ops} \addlegendentry{OPS}
      \addlegendimage{cwt} \addlegendentry{CWT}
      \addlegendimage{sid} \addlegendentry{SID}
      \addlegendimage{fro} \addlegendentry{FRO}
    \end{axis}
  \end{tikzpicture}
  \vspace{-0.8em}


  \begin{subfigure}[b]{0.235\textwidth}
    \centering
    \begin{tikzpicture}[baseline=(current bounding box.south)]
      \begin{axis}[
        xlabel={Number Scale \(N\)},
        ylabel={ASR (\%)},
        ymin=99, ymax=100.2,
        xtick={0.0,0.64,1.21,1.78,2.44,3.6},
        xticklabels={10,17,21,26,37,101},
        ytick={99,99.5,100},
        grid=both,
        axis y line*=left,
        axis x line*=bottom,
      ]
        \addplot[sia] coordinates {(0.0,100.0) (0.64,100.0) (1.21,100.0) (1.78,100.0) (2.44,100.0) (3.6,100.0)};
        \addplot[decowa] coordinates {(0.0,99.7) (0.64,99.9) (1.21,99.9) (1.78,99.9) (2.44,100.0) (3.6,100.0)};
        \addplot[ops] coordinates {(0.0,99.5) (0.64,99.9) (1.21,99.9) (1.78,99.9) (2.44,100.0) (3.6,100.0)};
        \addplot[cwt] coordinates {(0.0,100.0) (0.64,100.0) (1.21,100.0) (1.78,100.0) (2.44,100.0) (3.6,100.0)};
        \addplot[sid] coordinates {(0.0,100.0) (0.64,100.0) (1.21,100.0) (1.78,100.0) (2.44,100.0) (3.6,100.0)};
        \addplot[fro] coordinates {(0.0,100.0) (0.64,100.0) (1.21,100.0) (1.78,100.0) (2.44,100.0) (3.6,100.0)};
      \end{axis}
    \end{tikzpicture}
    \caption{RN-18}
    \label{fig:numscale-rn18}
  \end{subfigure}
  \hfill
  \begin{subfigure}[b]{0.235\textwidth}
    \centering
    \begin{tikzpicture}[baseline=(current bounding box.south)]
      \begin{axis}[
        xlabel={Number Scale \(N\)},
        ylabel={ASR (\%)},
        ymin=75, ymax=100,
        xtick={0.0,0.64,1.21,1.78,2.44,3.6},
        xticklabels={10,17,21,26,37,101},
        ytick={75,85,95},
        grid=both,
        axis y line*=left,
        axis x line*=bottom,
      ]
        \addplot[sia] coordinates {(0.0,88.4) (0.64,90.9) (1.21,92.3) (1.78,92.3) (2.44,93.8) (3.6,94.9)};
        \addplot[decowa] coordinates {(0.0,80.4) (0.64,84.3) (1.21,85.8) (1.78,86.8) (2.44,88.2) (3.6,91.0)};
        \addplot[ops] coordinates {(0.0,79.8) (0.64,85.6) (1.21,87.5) (1.78,89.1) (2.44,90.2) (3.6,94.0)};
        \addplot[cwt] coordinates {(0.0,91.0) (0.64,92.8) (1.21,94.2) (1.78,94.3) (2.44,95.5) (3.6,96.6)};
        \addplot[sid] coordinates {(0.0,89.7) (0.64,92.2) (1.21,93.2) (1.78,93.9) (2.44,94.1) (3.6,96.2)};
        \addplot[fro] coordinates {(0.0,91.8) (0.64,94.7) (1.21,95.6) (1.78,95.8) (2.44,96.4) (3.6,97.6)};
      \end{axis}
    \end{tikzpicture}
    \caption{RN-50}
    \label{fig:numscale-rn50}
  \end{subfigure}
  \hfill
  \begin{subfigure}[b]{0.235\textwidth}
    \centering
    \begin{tikzpicture}[baseline=(current bounding box.south)]
      \begin{axis}[
        xlabel={Number Scale \(N\)},
        ylabel={ASR (\%)},
        ymin=70, ymax=100,
        xtick={0.0,0.64,1.21,1.78,2.44,3.6},
        xticklabels={10,17,21,26,37,101},
        ytick={70,80,90,100},
        grid=both,
        axis y line*=left,
        axis x line*=bottom,
      ]
        \addplot[sia] coordinates {(0.0,80.5) (0.64,86.1) (1.21,87.1) (1.78,88.5) (2.44,90.5) (3.6,92.3)};
        \addplot[decowa] coordinates {(0.0,74.8) (0.64,81.0) (1.21,82.0) (1.78,82.5) (2.44,85.0) (3.6,90.8)};
        \addplot[ops] coordinates {(0.0,75.2) (0.64,83.4) (1.21,84.6) (1.78,88.1) (2.44,89.5) (3.6,94.4)};
        \addplot[cwt] coordinates {(0.0,86.8) (0.64,89.6) (1.21,90.8) (1.78,90.9) (2.44,93.0) (3.6,94.0)};
        \addplot[sid] coordinates {(0.0,86.7) (0.64,90.4) (1.21,90.3) (1.78,92.0) (2.44,93.0) (3.6,94.8)};
        \addplot[fro] coordinates {(0.0,88.8) (0.64,92.1) (1.21,92.4) (1.78,93.3) (2.44,94.3) (3.6,96.4)};
      \end{axis}
    \end{tikzpicture}
    \caption{RN-101}
    \label{fig:numscale-rn101}
  \end{subfigure}
  \hfill
  \begin{subfigure}[b]{0.235\textwidth}
    \centering
    \begin{tikzpicture}[baseline=(current bounding box.south)]
      \begin{axis}[
        xlabel={Number Scale \(N\)},
        ylabel={ASR (\%)},
        ymin=90, ymax=100,
        xtick={0.0,0.64,1.21,1.78,2.44,3.6},
        xticklabels={10,17,21,26,37,101},
        ytick={90,95,100},
        grid=both,
        axis y line*=left,
        axis x line*=bottom,
      ]
        \addplot[sia] coordinates {(0.0,98.4) (0.64,99.1) (1.21,99.2) (1.78,99.4) (2.44,99.3) (3.6,99.4)};
        \addplot[decowa] coordinates {(0.0,95.5) (0.64,97.0) (1.21,98.0) (1.78,98.5) (2.44,98.9) (3.6,99.2)};
        \addplot[ops] coordinates {(0.0,95.2) (0.64,96.6) (1.21,96.7) (1.78,97.9) (2.44,98.1) (3.6,99.2)};
        \addplot[cwt] coordinates {(0.0,99.2) (0.64,99.3) (1.21,99.3) (1.78,99.5) (2.44,99.5) (3.6,99.6)};
        \addplot[sid] coordinates {(0.0,98.1) (0.64,98.7) (1.21,99.0) (1.78,99.0) (2.44,99.2) (3.6,99.4)};
        \addplot[fro] coordinates {(0.0,99.0) (0.64,99.4) (1.21,99.5) (1.78,99.5) (2.44,99.6) (3.6,99.8)};
      \end{axis}
    \end{tikzpicture}
    \caption{VGG-16}
    \label{fig:numscale-vgg16}
  \end{subfigure}

  \par\vspace{0.65em}


  \begin{subfigure}[b]{0.235\textwidth}
    \centering
    \begin{tikzpicture}[baseline=(current bounding box.south)]
      \begin{axis}[
        xlabel={Number Scale \(N\)},
        ylabel={ASR (\%)},
        ymin=75, ymax=100,
        xtick={0.0,0.64,1.21,1.78,2.44,3.6},
        xticklabels={10,17,21,26,37,101},
        ytick={75,85,95},
        grid=both,
        axis y line*=left,
        axis x line*=bottom,
      ]
        \addplot[sia] coordinates {(0.0,86.1) (0.64,89.7) (1.21,90.7) (1.78,91.8) (2.44,92.6) (3.6,94.7)};
        \addplot[decowa] coordinates {(0.0,78.7) (0.64,83.9) (1.21,84.9) (1.78,86.3) (2.44,88.5) (3.6,91.3)};
        \addplot[ops] coordinates {(0.0,79.9) (0.64,86.2) (1.21,87.6) (1.78,89.3) (2.44,91.2) (3.6,95.2)};
        \addplot[cwt] coordinates {(0.0,90.5) (0.64,92.7) (1.21,92.9) (1.78,93.8) (2.44,94.7) (3.6,96.2)};
        \addplot[sid] coordinates {(0.0,89.2) (0.64,92.0) (1.21,93.0) (1.78,94.3) (2.44,94.5) (3.6,96.1)};
        \addplot[fro] coordinates {(0.0,92.0) (0.64,93.2) (1.21,94.1) (1.78,95.1) (2.44,95.8) (3.6,96.8)};
      \end{axis}
    \end{tikzpicture}
    \caption{ResNeXt-50}
    \label{fig:numscale-regx50}
  \end{subfigure}
  \hfill
  \begin{subfigure}[b]{0.235\textwidth}
    \centering
    \begin{tikzpicture}[baseline=(current bounding box.south)]
      \begin{axis}[
        xlabel={Number Scale \(N\)},
        ylabel={ASR (\%)},
        ymin=85, ymax=100,
        xtick={0.0,0.64,1.21,1.78,2.44,3.6},
        xticklabels={10,17,21,26,37,101},
        ytick={85,90,95,100},
        grid=both,
        axis y line*=left,
        axis x line*=bottom,
      ]
        \addplot[sia] coordinates {(0.0,87.0) (0.64,89.6) (1.21,90.8) (1.78,91.2) (2.44,92.7) (3.6,93.7)};
        \addplot[decowa] coordinates {(0.0,86.7) (0.64,90.2) (1.21,90.6) (1.78,91.8) (2.44,93.4) (3.6,94.8)};
        \addplot[ops] coordinates {(0.0,88.7) (0.64,93.5) (1.21,94.7) (1.78,95.5) (2.44,96.2) (3.6,98.4)};
        \addplot[cwt] coordinates {(0.0,94.4) (0.64,94.8) (1.21,95.2) (1.78,95.5) (2.44,96.5) (3.6,96.6)};
        \addplot[sid] coordinates {(0.0,93.2) (0.64,95.7) (1.21,96.2) (1.78,96.4) (2.44,96.2) (3.6,97.6)};
        \addplot[fro] coordinates {(0.0,94.1) (0.64,96.7) (1.21,96.3) (1.78,97.0) (2.44,97.6) (3.6,98.3)};
      \end{axis}
    \end{tikzpicture}
    \caption{Inception-v3}
    \label{fig:numscale-incv3}
  \end{subfigure}
  \hfill
  \begin{subfigure}[b]{0.235\textwidth}
    \centering
    \begin{tikzpicture}[baseline=(current bounding box.south)]
      \begin{axis}[
        xlabel={Number Scale \(N\)},
        ylabel={ASR (\%)},
        ymin=85, ymax=100,
        xtick={0.0,0.64,1.21,1.78,2.44,3.6},
        xticklabels={10,17,21,26,37,101},
        ytick={85,90,95,100},
        grid=both,
        axis y line*=left,
        axis x line*=bottom,
      ]
        \addplot[sia] coordinates {(0.0,89.6) (0.64,92.9) (1.21,93.3) (1.78,93.1) (2.44,94.2) (3.6,94.8)};
        \addplot[decowa] coordinates {(0.0,87.5) (0.64,90.4) (1.21,91.4) (1.78,93.5) (2.44,93.6) (3.6,96.2)};
        \addplot[ops] coordinates {(0.0,88.8) (0.64,93.6) (1.21,94.8) (1.78,95.4) (2.44,96.2) (3.6,97.8)};
        \addplot[cwt] coordinates {(0.0,93.5) (0.64,95.2) (1.21,95.6) (1.78,96.2) (2.44,96.5) (3.6,97.5)};
        \addplot[sid] coordinates {(0.0,94.5) (0.64,96.3) (1.21,96.9) (1.78,97.3) (2.44,98.2) (3.6,98.4)};
        \addplot[fro] coordinates {(0.0,95.2) (0.64,96.7) (1.21,96.8) (1.78,97.1) (2.44,97.3) (3.6,98.1)};
      \end{axis}
    \end{tikzpicture}
    \caption{Inception-v4}
    \label{fig:numscale-incv4}
  \end{subfigure}
  \hfill
  \begin{subfigure}[b]{0.235\textwidth}
    \centering
    \begin{tikzpicture}[baseline=(current bounding box.south)]
      \begin{axis}[
        xlabel={Number Scale \(N\)},
        ylabel={ASR (\%)},
        ymin=35, ymax=85,
        xtick={0.0,0.64,1.21,1.78,2.44,3.6},
        xticklabels={10,17,21,26,37,101},
        ytick={35,50,65,80},
        grid=both,
        axis y line*=left,
        axis x line*=bottom,
      ]
        \addplot[sia] coordinates {(0.0,40.0) (0.64,45.2) (1.21,47.0) (1.78,48.3) (2.44,50.5) (3.6,51.9)};
        \addplot[decowa] coordinates {(0.0,47.3) (0.64,53.6) (1.21,53.7) (1.78,56.4) (2.44,59.0) (3.6,65.3)};
        \addplot[ops] coordinates {(0.0,48.0) (0.64,55.5) (1.21,58.3) (1.78,59.9) (2.44,64.9) (3.6,75.3)};
        \addplot[cwt] coordinates {(0.0,53.0) (0.64,57.7) (1.21,58.7) (1.78,59.3) (2.44,61.6) (3.6,65.4)};
        \addplot[sid] coordinates {(0.0,55.6) (0.64,60.3) (1.21,62.2) (1.78,65.2) (2.44,66.3) (3.6,71.3)};
        \addplot[fro] coordinates {(0.0,63.0) (0.64,66.1) (1.21,69.1) (1.78,70.4) (2.44,71.6) (3.6,76.1)};
      \end{axis}
    \end{tikzpicture}
    \caption{ViT}
    \label{fig:numscale-vit}
  \end{subfigure}

  \par\vspace{0.65em}


  \begin{subfigure}[b]{0.235\textwidth}
    \centering
    \begin{tikzpicture}[baseline=(current bounding box.south)]
      \begin{axis}[
        xlabel={Number Scale \(N\)},
        ylabel={ASR (\%)},
        ymin=45, ymax=85,
        xtick={0.0,0.64,1.21,1.78,2.44,3.6},
        xticklabels={10,17,21,26,37,101},
        ytick={45,60,75},
        grid=both,
        axis y line*=left,
        axis x line*=bottom,
      ]
        \addplot[sia] coordinates {(0.0,52.2) (0.64,57.8) (1.21,58.7) (1.78,58.9) (2.44,62.4) (3.6,65.4)};
        \addplot[decowa] coordinates {(0.0,52.0) (0.64,59.2) (1.21,60.0) (1.78,62.6) (2.44,66.2) (3.6,70.0)};
        \addplot[ops] coordinates {(0.0,51.6) (0.64,59.9) (1.21,62.1) (1.78,64.7) (2.44,68.1) (3.6,77.1)};
        \addplot[cwt] coordinates {(0.0,61.2) (0.64,68.1) (1.21,68.2) (1.78,70.4) (2.44,71.5) (3.6,74.7)};
        \addplot[sid] coordinates {(0.0,63.0) (0.64,66.6) (1.21,68.2) (1.78,69.6) (2.44,71.4) (3.6,74.8)};
        \addplot[fro] coordinates {(0.0,71.4) (0.64,76.3) (1.21,76.4) (1.78,77.2) (2.44,79.9) (3.6,82.9)};
      \end{axis}
    \end{tikzpicture}
    \caption{PiT}
    \label{fig:numscale-pit}
  \end{subfigure}
  \hfill
  \begin{subfigure}[b]{0.235\textwidth}
    \centering
    \begin{tikzpicture}[baseline=(current bounding box.south)]
      \begin{axis}[
        xlabel={Number Scale \(N\)},
        ylabel={ASR (\%)},
        ymin=63, ymax=95,
        xtick={0.0,0.64,1.21,1.78,2.44,3.6},
        xticklabels={10,17,21,26,37,101},
        ytick={60,75,90},
        grid=both,
        axis y line*=left,
        axis x line*=bottom,
      ]
        \addplot[sia] coordinates {(0.0,72.2) (0.64,75.8) (1.21,76.5) (1.78,78.6) (2.44,80.2) (3.6,83.8)};
        \addplot[decowa] coordinates {(0.0,68.6) (0.64,73.6) (1.21,75.0) (1.78,77.5) (2.44,80.0) (3.6,84.8)};
        \addplot[ops] coordinates {(0.0,65.3) (0.64,73.0) (1.21,75.9) (1.78,79.6) (2.44,82.9) (3.6,88.7)};
        \addplot[cwt] coordinates {(0.0,79.4) (0.64,83.2) (1.21,84.9) (1.78,85.1) (2.44,87.3) (3.6,89.3)};
        \addplot[sid] coordinates {(0.0,78.8) (0.64,84.1) (1.21,83.7) (1.78,85.6) (2.44,86.9) (3.6,89.9)};
        \addplot[fro] coordinates {(0.0,84.3) (0.64,88.8) (1.21,89.9) (1.78,90.3) (2.44,91.4) (3.6,93.4)};
      \end{axis}
    \end{tikzpicture}
    \caption{Visformer}
    \label{fig:numscale-visformer}
  \end{subfigure}
  \hfill
  \begin{subfigure}[b]{0.235\textwidth}
    \centering
    \begin{tikzpicture}[baseline=(current bounding box.south)]
      \begin{axis}[
        xlabel={Number Scale \(N\)},
        ylabel={ASR (\%)},
        ymin=65, ymax=96,
        xtick={0.0,0.64,1.21,1.78,2.44,3.6},
        xticklabels={10,17,21,26,37,101},
        ytick={65,80,95},
        grid=both,
        axis y line*=left,
        axis x line*=bottom,
      ]
        \addplot[sia] coordinates {(0.0,73.4) (0.64,76.5) (1.21,77.1) (1.78,79.4) (2.44,80.7) (3.6,84.3)};
        \addplot[decowa] coordinates {(0.0,70.8) (0.64,74.9) (1.21,76.5) (1.78,79.7) (2.44,78.7) (3.6,84.6)};
        \addplot[ops] coordinates {(0.0,70.0) (0.64,75.8) (1.21,79.0) (1.78,80.7) (2.44,84.8) (3.6,88.3)};
        \addplot[cwt] coordinates {(0.0,79.7) (0.64,83.5) (1.21,83.0) (1.78,84.3) (2.44,87.2) (3.6,88.1)};
        \addplot[sid] coordinates {(0.0,80.3) (0.64,83.1) (1.21,83.8) (1.78,85.6) (2.44,87.2) (3.6,89.3)};
        \addplot[fro] coordinates {(0.0,84.7) (0.64,88.2) (1.21,90.0) (1.78,90.4) (2.44,91.8) (3.6,95.0)};
      \end{axis}
    \end{tikzpicture}
    \caption{Swin}
    \label{fig:numscale-swin}
  \end{subfigure}
  \hfill
  \begin{subfigure}[b]{0.235\textwidth}
    \centering
    \begin{tikzpicture}[baseline=(current bounding box.south)]
      \begin{axis}[
        xlabel={Number Scale \(N\)},
        ylabel={ASR (\%)},
        ymin=45, ymax=90,
        xtick={0.0,0.64,1.21,1.78,2.44,3.6},
        xticklabels={10,17,21,26,37,101},
        ytick={45,60,75,90},
        grid=both,
        axis y line*=left,
        axis x line*=bottom,
      ]
        \addplot[sia] coordinates {(0.0,54.0) (0.64,57.5) (1.21,59.8) (1.78,61.6) (2.44,63.0) (3.6,66.2)};
        \addplot[decowa] coordinates {(0.0,56.3) (0.64,62.6) (1.21,64.1) (1.78,65.3) (2.44,68.8) (3.6,72.7)};
        \addplot[ops] coordinates {(0.0,55.0) (0.64,64.4) (1.21,67.4) (1.78,70.9) (2.44,74.0) (3.6,82.6)};
        \addplot[cwt] coordinates {(0.0,65.1) (0.64,70.3) (1.21,71.0) (1.78,72.4) (2.44,74.2) (3.6,77.9)};
        \addplot[sid] coordinates {(0.0,65.9) (0.64,71.8) (1.21,72.0) (1.78,73.9) (2.44,75.8) (3.6,79.2)};
        \addplot[fro] coordinates {(0.0,73.5) (0.64,77.6) (1.21,78.9) (1.78,80.4) (2.44,82.1) (3.6,85.0)};
      \end{axis}
    \end{tikzpicture}
    \caption{CaiT}
    \label{fig:numscale-cait}
  \end{subfigure}

  \caption{Attack success rates (\%) of different input transformation-based attacks under different number scales \(N\), using ResNet-18 as the white-box source model. The results are evaluated on twelve black-box target models, including both CNN and Transformer-based architectures.}
  \label{fig:numscale}
\end{figure*}

As shown in \cref{fig:numscale}, increasing \(N\) generally improves the attack success rates of input transformation-based attacks, indicating that a larger implicit ensemble can provide more stable and transferable update directions. However, the performance gain differs across methods and target architectures. Some methods benefit more from large \(N\), while others show limited improvement when additional transformed views become redundant or less informative.

Across different number scales, FRO consistently shows strong transferability. In particular, FRO remains competitive under relatively small ensemble scales such as \(N=10\), \(17\), \(21\), \(26\), and \(37\), and maintains strong performance when \(N\) increases to \(101\). This indicates that the Local Scaling Operator and Projection Operator can induce effective and complementary frontend-response variations without relying only on a large number of sampled views. These observations also justify using a unified \(N\) in the main comparison, so that performance differences can be more fairly attributed to transformation design rather than ensemble scale.\textbf{}
\subsection{Transferability under Different Source Models}
\label{sec:source_model_transferability}

Table~\ref{tab:model_attack} reports the transferability of different attacks under five source-model settings. We use ResNet-18, Inception-v3, ViT-B/16, ResNet-101, and DenseNet-121 as white-box source models, and evaluate the generated adversarial examples on six CNN and five Transformer-based black-box target models. For all input transformation-based attacks, the implicit ensemble size is fixed to \(N=21\).

\begin{table*}[!t]
\centering
\caption{
Attack success rates (\%) of different attack methods on six CNNs and five ViT-based models
under five source-model settings. 
For each source-model block, the best result in each target-model column is highlighted in bold. 
$^*$ denotes the white-box attack result, where the source model and target model are the same. 
$^\dagger$ denotes the best black-box result in each target-model column across all source-model blocks,
excluding white-box results.
}
\resizebox{\textwidth}{!}{%
\begin{tabular}{c|ccccccccccc}
\hline
\multirow{2}{*}{Attack} & \multicolumn{11}{c}{ResNet-18 $\Rightarrow$} \\
 & ResNet-18 & ResNet-50 & ResNext-50 & DenseNet-121 & Inception-v3 & Inception-v4 & ViT & PiT & Visformer & Swin & CaiT \\ \hline
SIA & $\mathbf{100}^{*}$ & 92.3 & 90.7 & 99.1 & 90.8 & 93.3 & 47 & 58.7 & 76.5 & 77.1 & 59.8 \\
DeCoWA & $99.9^{*}$ & 85.8 & 84.9 & 97.7 & 90.6 & 91.4 & 53.7 & 60 & 75 & 76.5 & 64.1 \\
OPS & $99.9^{*}$ & 87.5 & 87.6 & 98.5 & 94.7 & 94.8 & 58.3 & 62.1 & 75.9 & 79 & 67.4 \\
CWT & $\mathbf{100}^{*}$ & 93.2 & 93 & 99.2 & 96.2 & $\mathbf{96.9}$ & 62.2 & 68.2 & 83.7 & 83.8 & 72 \\
SID & $\mathbf{100}^{*}$ & 89 & 87.3 & 99.1 & 90.2 & 92.3 & 46.4 & 59.4 & 74.9 & 76.9 & 56.4 \\
FRO & $\mathbf{100}^{*}$ & $\mathbf{95.6}$ & $\mathbf{94.1}$ & $\mathbf{99.6}^{\dagger}$ & $\mathbf{96.3}$ & 96.8 & $\mathbf{69.1}$ & $\mathbf{76.4}$ & $\mathbf{89.9}$ & $\mathbf{90}$ & $\mathbf{78.9}$ \\ \hline

\multirow{2}{*}{Attack} & \multicolumn{11}{c}{Inception-v3 $\Rightarrow$} \\
 & ResNet-18 & ResNet-50 & ResNext-50 & DenseNet-121 & Inception-v3 & Inception-v4 & ViT & PiT & Visformer & Swin & CaiT \\ \hline
SIA & 93 & 85.5 & 85.3 & 95 & $99.8^{*}$ & 94.8 & 48 & 63.8 & 75.9 & 76.7 & 62 \\
DeCoWA & 87.7 & 75.4 & 74.7 & 90.6 & $99.3^{*}$ & 91.5 & 43.7 & 51.2 & 63.3 & 67.2 & 52.6 \\
OPS & 88.5 & 73.2 & 75.3 & 89.9 & $99.2^{*}$ & 93.5 & 42.8 & 48.7 & 62 & 63.9 & 52.5 \\
CWT & 89.8 & 82.2 & 83 & 93.6 & $\mathbf{100}^{*}$ & 94.4 & 48.6 & 58.7 & 72.6 & 73 & 59.1 \\
SID & 91 & 78.9 & 80.2 & 93.6 & $99.7^{*}$ & 94.8 & 48.4 & 56.2 & 70.1 & 72.5 & 58.4 \\
FRO & $\mathbf{94.4}$ & $\mathbf{87.6}$ & $\mathbf{88.7}$ & $\mathbf{95.9}$ & $99.5^{*}$ & $\mathbf{96.4}$ & $\mathbf{61}$ & $\mathbf{71.4}$ & $\mathbf{82.1}$ & $\mathbf{83.1}$ & $\mathbf{73.2}$ \\ \hline

\multirow{2}{*}{Attack} & \multicolumn{11}{c}{ViT $\Rightarrow$} \\
 & ResNet-18 & ResNet-50 & ResNext-50 & DenseNet-121 & Inception-v3 & Inception-v4 & ViT & PiT & Visformer & Swin & CaiT \\ \hline
SIA & 81.8 & 79.4 & 78.8 & 83.4 & 78.1 & 79.4 & $\mathbf{99.7}^{*}$ & 85.8 & 87.4 & $\mathbf{90.2}^{\dagger}$ & $\mathbf{92.8}^{\dagger}$ \\
DeCoWA & 82.6 & 71.7 & 73 & 83.7 & 76.3 & 80.2 & $95.9^{*}$ & 79.9 & 79.9 & 80.8 & 79 \\
OPS & 85 & 77.9 & 78.4 & 85.1 & $\mathbf{84.9}$ & 85.3 & $93.9^{*}$ & 81.1 & 82.6 & 84.1 & 85 \\
CWT & 84.2 & 81.3 & $\mathbf{82.8}$ & 86 & 83 & 84.5 & $99.1^{*}$ & $\mathbf{89.7}^{\dagger}$ & 87.7 & 88.9 & 90.2 \\
SID & $\mathbf{85.7}$ & $\mathbf{82.4}$ & 82.1 & 88 & 84.5 & $\mathbf{86.3}$ & $98.8^{*}$ & 89.6 & $\mathbf{88.9}$ & 89 & 90.3 \\
FRO & 85.1 & 80.4 & 80.9 & $\mathbf{88.4}$ & 84 & $\mathbf{86.3}$ & $98.6^{*}$ & 89.3 & 87.8 & 89.2 & 87.5 \\ \hline

\multirow{2}{*}{Attack} & \multicolumn{11}{c}{ResNet-101 $\Rightarrow$} \\
 & ResNet-18 & ResNet-50 & ResNext-50 & DenseNet-121 & Inception-v3 & Inception-v4 & ViT & PiT & Visformer & Swin & CaiT \\ \hline
SIA & 87.1 & 89.9 & 89.5 & 89.2 & 79.5 & 82.4 & 51.9 & 71.8 & 78.6 & 76.2 & 66.1 \\
DeCoWA & 91.2 & 90.3 & 89.3 & 93.9 & 87.3 & 89.6 & 62.1 & 74.4 & 82.3 & 80.6 & 72.9 \\
OPS & 91.7 & 89.5 & 90.1 & 92.8 & $\mathbf{90.5}$ & 90.9 & $\mathbf{72.9}^{\dagger}$ & 79.4 & 85.2 & 83.8 & $\mathbf{80.8}$ \\
CWT & 87.6 & 91.1 & 91.1 & 91.8 & 85.7 & 88 & 64.6 & 77.9 & 85.1 & 83.2 & 76.4 \\
SID & 91.2 & 90.8 & 91.2 & 92.3 & 89.1 & 90.2 & 71.3 & 81.2 & 86.7 & 85.7 & 80.6 \\
FRO & $\mathbf{92.9}$ & $\mathbf{92.0}$ & $\mathbf{91.5}$ & $\mathbf{94.2}$ & 90.2 & $\mathbf{91.8}$ & 71.6 & $\mathbf{81.6}$ & $\mathbf{87.3}$ & $\mathbf{86.5}$ & 80.7 \\ \hline

\multirow{2}{*}{Attack} & \multicolumn{11}{c}{DenseNet-121 $\Rightarrow$} \\
 & ResNet-18 & ResNet-50 & ResNext-50 & DenseNet-121 & Inception-v3 & Inception-v4 & ViT & PiT & Visformer & Swin & CaiT \\ \hline
SIA & 98.9 & 96 & 95.6 & $\mathbf{100}^{*}$ & 93 & 94.2 & 53.3 & 67 & 84.6 & 82.6 & 66.3 \\
DeCoWA & 97.4 & 89.3 & 88.3 & $99.9^{*}$ & 91 & 93.9 & 55.7 & 64.1 & 78.3 & 78.8 & 66.8 \\
OPS & 98.3 & 92.4 & 92.2 & $99.7^{*}$ & 96 & 96.7 & 64.1 & 70 & 83 & 82.6 & 73.4 \\
CWT & 98.4 & 94 & 94.6 & $\mathbf{100}^{*}$ & 94.6 & 95.6 & 59.8 & 69.8 & 86.6 & 83.7 & 70.3 \\
SID & $\mathbf{99}^{\dagger}$ & 95.5 & 95.8 & $\mathbf{100}^{*}$ & 96.3 & 97 & 65.2 & 73.6 & 88.8 & 85.8 & 75.6 \\
FRO & 98.8 & $\mathbf{96.3}^{\dagger}$ & $\mathbf{96.6}^{\dagger}$ & $\mathbf{100}^{*}$ & $\mathbf{96.6}^{\dagger}$ & $\mathbf{97.2}^{\dagger}$ & $\mathbf{70.1}$ & $\mathbf{79.8}$ & $\mathbf{90.3}^{\dagger}$ & $\mathbf{89.4}$ & $\mathbf{80.8}$ \\ \hline
\end{tabular}%
}
\label{tab:model_attack}
\end{table*}

FRO achieves stable transferability across different source models. In the block-wise comparison, FRO obtains 31 best black-box results, more than SID, OPS, CWT, SIA, and DeCoWA. This suggests that FRO is not tailored to a specific surrogate architecture, but remains effective when the white-box source model changes.

The results also reveal that using a deeper or more complex source model does not always lead to better transferability. For example, using ResNet-18 as the source model gives higher FRO results than using ResNet-101 on several target models, including ResNet-50, ResNeXt-50, DenseNet-121, Inception-v3, Inception-v4, Visformer, and Swin. This is consistent with our motivation: in input transformation-based attacks, transferability depends not only on source-model capacity, but also on whether the source frontend preserves transformation-induced response differences that contribute to gradient aggregation.

We also report the global black-box best result for each target model. This metric selects the highest black-box success rate across different source-model settings while excluding the white-box case, reflecting a practical scenario where an attacker may choose among multiple surrogate models. Under this criterion, FRO achieves the best result on 6 out of 11 target models, including ResNet-50, ResNeXt-50, DenseNet-121, Inception-v3, Inception-v4, and Visformer.

Even when FRO is not the best in a target column, its gap to the best result is often small. For example, the best result on ResNet-18 is \(99.0\%\), while FRO reaches \(98.8\%\). On PiT and Swin, FRO is only \(0.4\%\) and \(0.8\%\) lower than the best results, respectively. Overall, these results show that FRO is robust to source-model selection and achieves the best overall balance across block-wise best results, global black-box best results, and average black-box success rate.

\subsection{Defense Experiments}
\label{sec:defense_experiments}

Following the transferability experiments, we further evaluate FRO against defended models, including three ensemble adversarially trained models, RS, HGD, NRP-based settings, AT, and DiffPure. All adversarial examples are generated on Inception-v3 using MI-FGSM as the base attack, with the transformation sampling budget fixed at \(N=21\). The results are reported in \cref{tab:defense}.

\begin{table*}[!t]
\centering
\caption{
	Attack success rates (\%) of different attack methods on eight image classifiers with different defenses, using Inception-v3 as the
	white-box model and setting the number scale to 21. The best results are highlighted in bold.
}
\resizebox{\textwidth}{!}{%
\begin{tabular}{c|llllllllc}
\hline
\multirow{2}{*}{Attack} & \multicolumn{8}{c}{Inception-v3  $\Rightarrow$} \\
 & \multicolumn{1}{c}{\begin{tabular}[c]{@{}c@{}}Inception-v3\\ (ens3)\end{tabular}} & \multicolumn{1}{c}{\begin{tabular}[c]{@{}c@{}}Inception-v3\\ (ens4)\end{tabular}} & \multicolumn{1}{c}{\begin{tabular}[c]{@{}c@{}}Inception-ResNet-v2\\ (ens2)\end{tabular}}& \multicolumn{1}{c}{RS} & \multicolumn{1}{c}{HGD} & \multicolumn{1}{c}{\begin{tabular}[c]{@{}c@{}}NRP\\ ($\rightarrow$ResNet18)\end{tabular}} & \multicolumn{1}{c}{\begin{tabular}[c]{@{}c@{}}NRP\\ ($\rightarrow$ViT)\end{tabular}} & AT & DiffPure\\ \hline
SIA & 90.9 & 89.4 & 80.4 & 27.4 & 89.4 & 76.1 & 30.1 & 37.8 & 28.6\\
DeCoWA & 85.5 & 85.1 & 76.1 & 32.6 & 79.3 & 74.9 & 29.9 & 40.2 & 39.6\\
OPS & 86.5 & 88.1 & 79.7 & \textbf{36} & 82.5 & 78.9 & 33.7 & \textbf{41.6} & 31.6\\
CWT & 92.6 & 89.8 & 84.2 & 29.5 & 89.4 & 76.1 & 31 & 39.1 & 39.5\\
SID & 91.7 & 91.6 & 82.9 & 32.80 & 85.9 & 79.2 & 34.3 & 40 & 45.9\\
FRO & \textbf{96.2} & \textbf{93.3} & \textbf{89.7} & 35.8 & \textbf{92.3} & \textbf{81.7} & \textbf{41.4} & \textbf{43.0} & \textbf{47.6}\\
\hline
\end{tabular}%
}\label{tab:defense}
\end{table*}

FRO achieves the best attack success rate in eight of the nine defended settings, demonstrating strong transferability across substantially different defense strategies. In particular, FRO obtains success rates of \(96.2\%\), \(93.3\%\), and \(89.7\%\) against the three ensemble adversarially trained models, outperforming all competing transformation-based attacks. It also achieves the highest success rates against HGD, both NRP-based settings, AT, and DiffPure. Against RS, FRO reaches \(35.8\%\), only \(0.2\) percentage points below the best result of \(36.0\%\) achieved by OPS.

Averaged over all nine defended settings, FRO achieves an attack success rate of \(69.0\%\), compared with \(64.92\%\) for the second-best method, SID. The consistent improvements across adversarial training, reconstruction-based preprocessing, purification, and randomized defense settings indicate that the adversarial examples generated by FRO are not specialized to a particular defense mechanism. Instead, they maintain strong cross-model and cross-defense transferability, further supporting the effectiveness of constructing transformations that induce transferable frontend-response variations across heterogeneous architectures.

\subsection{Ablation Experiments}
\label{sec:ablation_experiments}

We analyze the hyperparameters of FRO using ResNet-18 as the source model. All ablation experiments follow the same attack settings, with the number of transformed views fixed to \(N=30\) to reduce the variance caused by stochastic transformation sampling. After parameter selection, the selected values are fixed for subsequent experiments, while the main comparison follows the unified ensemble size of \(N=21\).

\paragraph{Local Scaling Operator.}
We first tune the Local Scaling Operator, including the minimum boundary distance \(d_b\), the minimum distance between segmentation points \(d_p\), the range-to-center ratio \(r\), and the number of segmentation points \(M\). The results in \cref{fig:pse-sub1,fig:pse-sub2,fig:pse-sub3,fig:pse-sub4} show that \(d_b=35\), \(d_p=40\), \(r=1\), and \(M=2\) provide the best overall transfer performance. These values are therefore used as the default Local Scaling parameters.

\paragraph{Projection Operator.}
We then tune the Projection Operator, which is controlled by the virtual focal length \(f_p\) and the maximum tilt angle \(A_{\max}\). As shown in \cref{fig:acss-focal-length}, \(f_p=1.5\) achieves the best overall performance and is selected as the default focal length. For the tilt angle, \(A_{\max}=30\) performs favorably when the Projection Operator is evaluated alone. However, in the final FRO setting, Projection is combined with Local Scaling, and an overly large projection angle may introduce excessive geometric distortion. Therefore, we further evaluate the combined setting and select \(A_{\max}=15\) as the default value.

\begin{figure*}[!t]
  \centering
  \definecolor{cnncolor}{HTML}{4477AA} 
\definecolor{vitcolor}{HTML}{EE7733} 
  \pgfplotsset{
    compat=1.18,
    every axis/.append style={
      height=3.8cm,
      width=\linewidth,
      xlabel style={font=\scriptsize},
      ylabel style={font=\scriptsize},
      tick label style={font=\scriptsize},
      grid style={dashed,gray!60},
      axis line style={draw=none},
      tick align=outside,
      enlargelimits=false,
      clip=false,
    },
    cnn/.style={mark=*, mark size=1.8pt, color=cnncolor, line width=1pt, smooth},
    vit/.style={mark=square*, mark size=1.8pt, color=vitcolor, line width=1pt, smooth},
  }

  \begin{subfigure}{\textwidth}
    \centering
    \begin{tikzpicture}
      \begin{axis}[
        hide axis,
        xmin=0, xmax=1, ymin=0, ymax=1,
        legend style={
          at={(0.5,1.0)},
          anchor=north,
          draw=none,
          fill=none,
          font=\scriptsize,
          legend columns=2,
          column sep=12pt,
        },
      ]
        \addlegendimage{cnn} \addlegendentry{CNNs models avg}
        \addlegendimage{vit} \addlegendentry{ViTs models avg}
      \end{axis}
    \end{tikzpicture}
  \end{subfigure}
  \\[-0.7em]  

  \begin{subfigure}[b]{0.22\textwidth}
    \centering

    \begin{tikzpicture}
      \begin{axis}[
        name=leftax,
        xlabel=Min distance between points,
        ylabel={CNNs avg attack success (\%)},
        ymin=95, ymax=98,
        ytick={96,97,98},
        xtick={0,10,20,30,40,50,60},
        grid=both,
        axis y line*=left,
        axis x line*=bottom,
      ]
        \addplot[cnn] table[col sep=comma, x=X, y=CNNs_avg]{data1.csv};
      \end{axis}
      \begin{axis}[
        at={(leftax.east)}, anchor=east, 
        ylabel={ViTs avg attack success (\%)},
        ymin=78, ymax=83,
        ytick={78,80,82},
        xtick=\empty,
        grid=both,
        axis y line*=right,
        axis x line=none,
      ]
        \addplot[vit] table[col sep=comma, x=X, y=Vits_avg]{data1.csv};
      \end{axis}
    \end{tikzpicture}
    \caption{Min distance between points}
    \label{fig:pse-sub1}
  \end{subfigure}
  \hfill
  \begin{subfigure}[b]{0.22\textwidth}
    \centering

    \begin{tikzpicture}
      \begin{axis}[
        xlabel=Min Distance to Image Edges,
        ylabel=CNNs avg attack success (\%),
        ymin=95, ymax=98,
        ytick={96,97,98},
        xtick={10,20,30,40,50,60,70,80,90},
        grid=both,
        axis y line*=left,
        axis x line*=bottom,
      ]
        \addplot[cnn] table[col sep=comma, x=X, y=CNNs_avg] {data2.csv};
      \end{axis}
      \begin{axis}[
        ymin=78, ymax=83,
        ytick={78,80,82},
        ylabel=Vits avg attack success (\%),
        xtick=\empty,
        grid=both,
        axis y line*=right,
        axis x line=none,
      ]
        \addplot[vit] table[col sep=comma, x=X, y=Vits_avg] {data2.csv};
      \end{axis}
    \end{tikzpicture}
    \caption{Min distance to image edges}
    \label{fig:pse-sub2}
  \end{subfigure}
  \hfill
  \begin{subfigure}[b]{0.22\textwidth}
    \centering

    \begin{tikzpicture}
      \begin{axis}[
        ylabel={CNNs avg attack success (\%)},
        xlabel=Uniform range-to-center ratio,
        ymin=96, ymax=98,
        ytick={96,97,98},
        xtick={0,0.4,0.8,1.2,1.6,2.0},
        xticklabels={0,0.4,0.8,1.2,1.6,2.0},
        grid=both,
        axis y line*=left,
        axis x line*=bottom,
      ]
        \addplot[cnn] table[col sep=comma, x=X, y=CNNs_avg] {data3.csv};
      \end{axis}
      \begin{axis}[
        ylabel=Vits avg attack success (\%),
        ymin=80, ymax=83,
        ytick={80,81,82},
        xtick=\empty,
        grid=both,
        axis y line*=right,
        axis x line=none,
      ]
        \addplot[vit] table[col sep=comma, x=X, y=Vits_avg] {data3.csv};
      \end{axis}
    \end{tikzpicture}
    \caption{Uniform range-to-center ratio}
    \label{fig:pse-sub3}
  \end{subfigure}
    \hfill
  \begin{subfigure}[b]{0.22\textwidth}
    \centering

    \begin{tikzpicture}
      \begin{axis}[
        ylabel={CNNs avg attack success (\%)},
        xlabel=Num segmentation points,
        ymin=95, ymax=98,
        ytick={95,96,97,98},
        xtick={1,2,3,4,5},
        grid=both,
        axis y line*=left,
        axis x line*=bottom,
      ]
        \addplot[cnn] table[col sep=comma, x=X, y=CNNs_avg] {data4.csv};
      \end{axis}
      \begin{axis}[
        ylabel=Vits avg attack success (\%),
        ymin=82, ymax=85,
        ytick={82,83,84,85},
        xtick=\empty,
        grid=both,
        axis y line*=right,
        axis x line=none,
      ]
        \addplot[vit] table[col sep=comma, x=X, y=Vits_avg] {data4.csv};
      \end{axis}
    \end{tikzpicture}
    \caption{Num segmentation points}
    \label{fig:pse-sub4}
  \end{subfigure}

    \caption{
Average attack success rates (\%) on six CNNs and five ViTs models under different parameter configurations, using ResNet-18 as the white-box model and setting the number scale to 30.
}
    \label{fig:parameter-selection-example}
\end{figure*}

\begin{figure*}[!t]
  \centering

  \definecolor{cnncolor}{HTML}{4477AA}
  \definecolor{vitcolor}{HTML}{EE7733}

  \pgfplotsset{
    compat=1.18,
    every axis/.append style={
      height=3.5cm,
      width=\linewidth,
      xlabel style={font=\scriptsize},
      ylabel style={font=\scriptsize},
      tick label style={font=\scriptsize},
      grid style={dashed,gray!60},
      axis line style={draw=none},
      tick align=outside,
      enlargelimits=false,
      clip=false,
    },
    cnn/.style={
      mark=*,
      mark size=1.8pt,
      color=cnncolor,
      line width=1pt,
      smooth
    },
    vit/.style={
      mark=square*,
      mark size=1.8pt,
      color=vitcolor,
      line width=1pt,
      smooth
    },
  }

  \begin{subfigure}{\textwidth}
    \centering
    \begin{tikzpicture}
      \begin{axis}[
        hide axis,
        xmin=0,
        xmax=1,
        ymin=0,
        ymax=1,
        legend style={
          at={(0.5,1.0)},
          anchor=north,
          draw=none,
          fill=none,
          font=\scriptsize,
          legend columns=2,
          column sep=12pt,
        },
      ]
        \addlegendimage{cnn}
        \addlegendentry{CNNs models avg}

        \addlegendimage{vit}
        \addlegendentry{ViTs models avg}
      \end{axis}
    \end{tikzpicture}
  \end{subfigure}

\vspace{-0.7em}

  \begin{subfigure}[b]{0.30\textwidth}
    \centering

    \begin{tikzpicture}
      \begin{axis}[
        xlabel={Focal length},
        ylabel={CNNs avg attack success (\%)},
        ymin=93.5,
        ymax=96.0,
        ytick={94,95,96},
        xtick={0.5,1.0,1.5,2.0,2.5,3.0,3.5},
        grid=both,
        axis y line*=left,
        axis x line*=bottom,
      ]
        \addplot[cnn]
        table[
          col sep=comma,
          x=X,
          y=CNNs_avg
        ]{acss_focal_length.csv};
      \end{axis}

      \begin{axis}[
        ylabel={ViTs avg attack success (\%)},
        ymin=68,
        ymax=74,
        ytick={68,70,72,74},
        xtick=\empty,
        grid=both,
        axis y line*=right,
        axis x line=none,
      ]
        \addplot[vit]
        table[
          col sep=comma,
          x=X,
          y=ViTs_avg
        ]{acss_focal_length.csv};
      \end{axis}
    \end{tikzpicture}

    \caption{Focal length}
    \label{fig:acss-focal-length}
  \end{subfigure}
  \hfill
  \begin{subfigure}[b]{0.30\textwidth}
    \centering

    \begin{tikzpicture}
      \begin{axis}[
        xlabel={Max angle},
        ylabel={CNNs avg attack success (\%)},
        ymin=91,
        ymax=96,
        ytick={91,92,93,94,95,96},
        xtick={5,10,15,20,25,30,35},
        grid=both,
        axis y line*=left,
        axis x line*=bottom,
      ]
        \addplot[cnn]
        table[
          col sep=comma,
          x=X,
          y=CNNs_avg
        ]{acss_max_angle.csv};
      \end{axis}

      \begin{axis}[
        ylabel={ViTs avg attack success (\%)},
        ymin=64,
        ymax=74,
        ytick={64,66,68,70,72,74},
        xtick=\empty,
        grid=both,
        axis y line*=right,
        axis x line=none,
      ]
        \addplot[vit]
        table[
          col sep=comma,
          x=X,
          y=ViTs_avg
        ]{acss_max_angle.csv};
      \end{axis}
    \end{tikzpicture}

    \caption{Max angle}
    \label{fig:acss-max-angle}
  \end{subfigure}
  \hfill
  \begin{subfigure}[b]{0.30\textwidth}
    \centering

    \begin{tikzpicture}
      \begin{axis}[
        xlabel={Max angle},
        ylabel={CNNs avg attack success (\%)},
        ymin=94,
        ymax=98,
        ytick={94,95,96,97,98},
        xtick={5,10,15,20,25,30,35},
        grid=both,
        axis y line*=left,
        axis x line*=bottom,
      ]
        \addplot[cnn]
        table[
          col sep=comma,
          x=X,
          y=CNNs_avg
        ]{acss_combined_max_angle.csv};
      \end{axis}

      \begin{axis}[
        ylabel={ViTs avg attack success (\%)},
        ymin=72,
        ymax=84,
        ytick={72,74,76,78,80,82,84},
        xtick=\empty,
        grid=both,
        axis y line*=right,
        axis x line=none,
      ]
        \addplot[vit]
        table[
          col sep=comma,
          x=X,
          y=ViTs_avg
        ]{acss_combined_max_angle.csv};
      \end{axis}
    \end{tikzpicture}

    \caption{Max angle in the combined setting}
    \label{fig:acss-combined-max-angle}
  \end{subfigure}

  \caption{
    Average attack success rates (\%) on CNN and ViT models under
    different focal length and max angle configurations.
    The combined setting evaluates \(A_{\max}\) after integrating
    the Projection Operator with the Local Scaling Operator.
  }
  \label{fig:acss-focal-angle-selection}
\end{figure*}

\section{Summary}

This paper revisits input transformation-based transfer attacks from the perspective of implicit ensemble construction and frontend response relations. By decomposing the source model into a frontend feature extractor and a backend classifier, input transformations are interpreted as a way to construct an implicit ensemble over transformation-induced frontend responses. This perspective clarifies how transformed views contribute to transferable adversarial perturbations.

Based on this perspective, we propose FRO, a Frontend Response-oriented input transformation attack. FRO combines a Local Scaling Operator and a Projection Operator to generate diverse yet semantically consistent transformed views. Experiments across different source models, target architectures, and defended models show that FRO achieves strong transferability and robustness. The frontend response relation analysis further shows that FRO reduces cross-architecture response-relation discrepancy between CNN and ViT frontends, supporting the effectiveness of designing input transformations from the frontend response perspective.

{
    \small
    \bibliographystyle{plain}
    \bibliography{main}
}
\appendix

\section{Implementation Details}
\label{app:implementation_details}

All experiments follow the dataset specification of \texttt{TransferAttack}. CWT is evaluated using the official implementation, while DeCoWA, SIA, SID, and OPS are adopted from the \texttt{TransferAttack} library without modification. We use the repository version available as of March~2025. Software versions, including \texttt{torchvision==0.13.1}, are reported for reproducibility.

For DiffPure, we use the official ImageNet implementation with DDPM-based purification. The input adversarial images are resized from \(224\times224\) to \(256\times256\) before purification and resized back to \(224\times224\) before classification. We set the diffusion noise scale to \(t=400\), use one sampling step, and adopt ResNet-101 as the downstream classifier. The evaluation is conducted on 1000 images with a batch size of 1 and a random seed of 1234.

For frontend response relation analysis, we use 1000 ImageNet validation images, with one image from each class. For each clean image, every method generates \(N=21\) transformed views. The same transformed views are fed into ResNet-18 and ViT-B/16, which serve as representative CNN and Transformer architectures. We extract frontend responses from \texttt{layer1.0.conv2} of ResNet-18 and \texttt{blocks.0} of ViT-B/16. For each image and method, we construct the CNN and ViT T-FRDMs defined in \cref{sec:frontend_gap}, and compute one Gap and one Corr value. Gap Mean and Corr Mean are obtained by averaging the corresponding values over all 1000 images. All methods use the same clean images, number of transformed views, and random seed.

\section{Full Results under Different Number Scales}
\label{app:number_scale_full}

This section provides the complete numerical results for the number-scale experiment discussed in Sec.~\ref{sec:number_scale_analysis}. We evaluate all input transformation-based attacks under the same set of transformed-view numbers, namely \(N \in \{10,17,21,26,37,101\}\). All methods use ResNet-18 as the white-box source model and are evaluated on twelve target models, including both CNN and Transformer-based architectures.

\begin{table*}[!t]
  \centering
  \caption{Attack success rates (\%) of different methods under different number scales.}
  \label{tab:numscale_full}

  \setlength{\tabcolsep}{1.2pt}
  \renewcommand{\arraystretch}{1.06}
  \resizebox{\textwidth}{!}{%
  \scriptsize
\begin{tabular}{c l| *{12}{c}}
\hline
  \multirow{2}{*}{N} & \multirow{2}{*}{method} & \multicolumn{12}{c}{ResNet-18 $\Rightarrow$} \\
  \cline{3-14}
&  & RN-18 & RN-50 & RN-101 & VGG-16 & RegX-50 & Inc-v3 & Inc-v4 & ViT & PiT & Vis. & Swin & CaiT   \\
 \hline

    \multirow{6}{*}{10}
    & SIA    & $\bm{100}$ & 88.4 & 80.5 & 98.4 & 86.1 & 87.0 & 89.6 & 40.0 & 52.2 & 72.2 & 73.4 & 54.0 \\
    & DeCoWA & 99.7 & 80.4 & 74.8 & 95.5 & 78.7 & 86.7 & 87.5 & 47.3 & 52.0 & 68.6 & 70.8 & 56.3 \\
    & OPS    & 99.5 & 79.8 & 75.2 & 95.2 & 79.9 & 88.7 & 88.8 & 48.0 & 51.6 & 65.3 & 70.0 & 55.0 \\
    & CWT    & $\bm{100}$ & 91.0 & 86.8 & 99.2 & 90.5 & 94.4 & 93.5 & 53.0 & 61.2 & 79.4 & 79.7 & 65.1 \\
    & SID    & $\bm{100}$ & 89.7 & 86.7 & 98.1 & 89.2 & 93.2 & 94.5 & 55.6 & 63.0 & 78.8 & 80.3 & 65.9 \\
    & FRO    & $\bm{100}$ & $\bm{91.8}$ & $\bm{88.8}$ & $\bm{99.0}$ & $\bm{92.0}$ & $\bm{94.1}$ & $\bm{95.2}$ & $\bm{63.0}$ & $\bm{71.4}$ & $\bm{84.3}$ & $\bm{84.7}$ & $\bm{73.5}$ \\
 \hline

    \multirow{6}{*}{17}
    & SIA    & $\bm{100}$ & 90.9 & 86.1 & 99.1 & 89.7 & 89.6 & 92.9 & 45.2 & 57.8 & 75.8 & 76.5 & 57.5 \\
    & DeCoWA & 99.9 & 84.3 & 81.0 & 97.0 & 83.9 & 90.2 & 90.4 & 53.6 & 59.2 & 73.6 & 74.9 & 62.6 \\
    & OPS    & 99.9 & 85.6 & 83.4 & 96.6 & 86.2 & 93.5 & 93.6 & 55.5 & 59.9 & 73.0 & 75.8 & 64.4 \\
    & CWT    & $\bm{100}$ & 92.8 & 89.6 & 99.3 & 92.7 & 94.8 & 95.2 & 57.7 & 68.1 & 83.2 & 83.5 & 70.3 \\
    & SID    & $\bm{100}$ & 92.2 & 90.4 & 98.7 & 92.0 & 95.7 & 96.3 & 60.3 & 66.6 & 84.1 & 83.1 & 71.8 \\
    & FRO    & $\bm{100}$ & $\bm{94.7}$ & $\bm{92.1}$ & $\bm{99.4}$ & $\bm{93.2}$ & $\bm{96.7}$ & $\bm{96.7}$ & $\bm{66.1}$ & $\bm{76.3}$ & $\bm{88.8}$ & $\bm{88.2}$ & $\bm{77.6}$ \\
\hline

    \multirow{6}{*}{21}
    & SIA    & $\bm{100}$ & 92.3 & 87.1 & 99.2 & 90.7 & 90.8 & 93.3 & 47.0 & 58.7 & 76.5 & 77.1 & 59.8 \\
    & DeCoWA & 99.9 & 85.8 & 82.0 & 98.0 & 84.9 & 90.6 & 91.4 & 53.7 & 60.0 & 75.0 & 76.5 & 64.1 \\
    & OPS    & 99.9 & 87.5 & 84.6 & 96.7 & 87.6 & 94.7 & 94.8 & 58.3 & 62.1 & 75.9 & 79.0 & 67.4 \\
    & CWT    & $\bm{100}$ & 94.2 & 90.8 & 99.3 & 92.9 & 95.2 & 95.6 & 58.7 & 68.2 & 84.9 & 83.0 & 71.0 \\
    & SID    & $\bm{100}$ & 93.2 & 90.3 & 99.0 & 93.0 & 96.2 & $\bm{96.9}$ & 62.2 & 68.2 & 83.7 & 83.8 & 72.0 \\
    & FRO    & $\bm{100}$ & $\bm{95.6}$ & $\bm{92.4}$ & $\bm{99.5}$ & $\bm{94.1}$ & $\bm{96.3}$ & 96.8 & $\bm{69.1}$ & $\bm{76.4}$ & $\bm{89.9}$ & $\bm{90.0}$ & $\bm{78.9}$ \\
\hline

    \multirow{6}{*}{26}
    & SIA    & $\bm{100}$ & 92.3 & 88.5 & 99.4 & 91.8 & 91.2 & 93.1 & 48.3 & 58.9 & 78.6 & 79.4 & 61.6 \\
    & DeCoWA & 99.9 & 86.8 & 82.5 & 98.5 & 86.3 & 91.8 & 93.5 & 56.4 & 62.6 & 77.5 & 79.7 & 65.3 \\
    & OPS    & 99.9 & 89.1 & 88.1 & 97.9 & 89.3 & 95.5 & 95.4 & 59.9 & 64.7 & 79.6 & 80.7 & 70.9 \\
    & CWT    & $\bm{100}$ & 94.3 & 90.9 & 99.5 & 93.8 & 95.5 & 96.2 & 59.3 & 70.4 & 85.1 & 84.3 & 72.4 \\
    & SID    & $\bm{100}$ & 93.9 & 92.0 & 99.0 & 94.3 & 96.4 & $\bm{97.3}$ & 65.2 & 69.6 & 85.6 & 85.6 & 73.9 \\
    & FRO    & $\bm{100}$ & $\bm{95.8}$ & $\bm{93.3}$ & $\bm{99.5}$ & $\bm{95.1}$ & $\bm{97.0}$ & 97.1 & $\bm{70.4}$ & $\bm{77.2}$ & $\bm{90.3}$ & $\bm{90.4}$ & $\bm{80.4}$ \\
\hline

    \multirow{6}{*}{37}
    & SIA    & $\bm{100}$ & 93.8 & 90.5 & 99.3 & 92.6 & 92.7 & 94.2 & 50.5 & 62.4 & 80.2 & 80.7 & 63.0 \\
    & DeCoWA & $\bm{100}$ & 88.2 & 85.0 & 98.9 & 88.5 & 93.4 & 93.6 & 59.0 & 66.2 & 80.0 & 78.7 & 68.8 \\
    & OPS    & $\bm{100}$ & 90.2 & 89.5 & 98.1 & 91.2 & 96.2 & 96.2 & 64.9 & 68.1 & 82.9 & 84.8 & 74.0 \\
    & CWT    & $\bm{100}$ & 95.5 & 93.0 & 99.5 & 94.7 & 96.5 & 96.5 & 61.6 & 71.5 & 87.3 & 87.2 & 74.2 \\
    & SID    & $\bm{100}$ & 94.1 & 93.0 & 99.2 & 94.5 & 96.2 & $\bm{98.2}$ & 66.3 & 71.4 & 86.9 & 87.2 & 75.8 \\
    & FRO    & $\bm{100}$ & $\bm{96.4}$ & $\bm{94.3}$ & $\bm{99.6}$ & $\bm{95.8}$ & $\bm{97.6}$ & 97.3 & $\bm{71.6}$ & $\bm{79.9}$ & $\bm{91.4}$ & $\bm{91.8}$ & $\bm{82.1}$ \\
\hline

    \multirow{6}{*}{101}
    & SIA    & $\bm{100}$ & 94.9 & 92.3 & 99.4 & 94.7 & 93.7 & 94.8 & 51.9 & 65.4 & 83.8 & 84.3 & 66.2 \\
    & DeCoWA & $\bm{100}$ & 91.0 & 90.8 & 99.2 & 91.3 & 94.8 & 96.2 & 65.3 & 70.0 & 84.8 & 84.6 & 72.7 \\
    & OPS    & $\bm{100}$ & 94.0 & 94.4 & 99.2 & 95.2 & $\bm{98.4}$ & 97.8 & 75.3 & 77.1 & 88.7 & 88.3 & 82.6 \\
    & CWT    & $\bm{100}$ & 96.6 & 94.0 & 99.6 & 96.2 & 96.6 & 97.5 & 65.4 & 74.7 & 89.3 & 88.1 & 77.9 \\
    & SID    & $\bm{100}$ & 96.2 & 94.8 & 99.4 & 96.1 & 97.6 & $\bm{98.4}$ & 71.3 & 74.8 & 89.9 & 89.3 & 79.2 \\
    & FRO    & $\bm{100}$ & $\bm{97.6}$ & $\bm{96.4}$ & $\bm{99.8}$ & $\bm{96.8}$ & 98.3 & 98.1 & $\bm{76.1}$ & $\bm{82.9}$ & $\bm{93.4}$ & $\bm{95.0}$ & $\bm{85.0}$ \\
    \bottomrule
  \end{tabular}
  }
\end{table*}

The complete table shows the detailed attack success rates corresponding to the trend visualization in \cref{fig:numscale}. Overall, increasing \(N\) tends to improve transferability, but the degree of improvement varies across different transformation methods. This supports the need to control the implicit ensemble size when comparing input transformation-based attacks. Under the same number scales, FRO consistently achieves strong performance across both CNN and Transformer-based target models, confirming the effectiveness and scalability of the proposed transformation operators.
\section{Frontend Response Relation Analysis}
\label{sec:exp_frontend_relation}

This experiment examines whether transformation-induced frontend response relations in a CNN source model are preserved in a heterogeneous ViT frontend. Following \cref{sec:frontend_gap}, we generate transformed views for 1000 ImageNet images and feed the same views into the frontend of ResNet-18 and an early block of ViT-B/16. We then construct the corresponding transformation-induced frontend response dissimilarity matrices and compare them using Gap and Corr. Gap Mean reflects the average discrepancy between CNN and ViT response relations, while Corr Mean measures whether the relative relation structure among different transformed-view pairs is preserved.

\begin{table}[t]
\centering
\caption{
CNN--ViT frontend response relation results over 1000 ImageNet images.
A lower Gap Mean and a higher Corr Mean indicate better performance.
}
\label{tab:frontend_gap_corr}
\small
\setlength{\tabcolsep}{8pt}
\begin{tabular}{lcc}
\toprule
Method & Gap Mean \(\downarrow\) & Corr Mean \(\uparrow\) \\
\midrule
FRO    & \textbf{0.0550} & 0.9173 \\
SID    & 0.1020 & 0.8559 \\
SIA    & 0.1158 & 0.8018 \\
OPS    & 0.1263 & \textbf{0.9280} \\
CWT    & 0.1340 & 0.9187 \\
DeCoWA & 0.1927 & 0.8703 \\
\bottomrule
\end{tabular}
\end{table}

FRO achieves the lowest Gap Mean of \(0.0550\). SID obtains the second-lowest Gap Mean of \(0.1020\). Compared with SID, FRO reduces the average frontend response-relation discrepancy between CNN and ViT by approximately \(46.1\%\). Meanwhile, FRO achieves a Corr Mean of \(0.9173\), which is close to the highest result of \(0.9280\) obtained by OPS.

The comparison between Gap and Corr further shows that these two metrics capture complementary aspects. OPS and CWT achieve high Corr Mean values of \(0.9280\) and \(0.9187\), respectively, but their Gap Mean values are \(0.1263\) and \(0.1340\), both substantially higher than that of FRO. SID exhibits a different behavior: although its Gap Mean is the second lowest, its Corr Mean is only \(0.8559\). Overall, FRO achieves the best balance between reducing relation-value discrepancy and preserving response-relation structure.

\begin{figure*}[t]
    \centering
    \includegraphics[width=\textwidth]{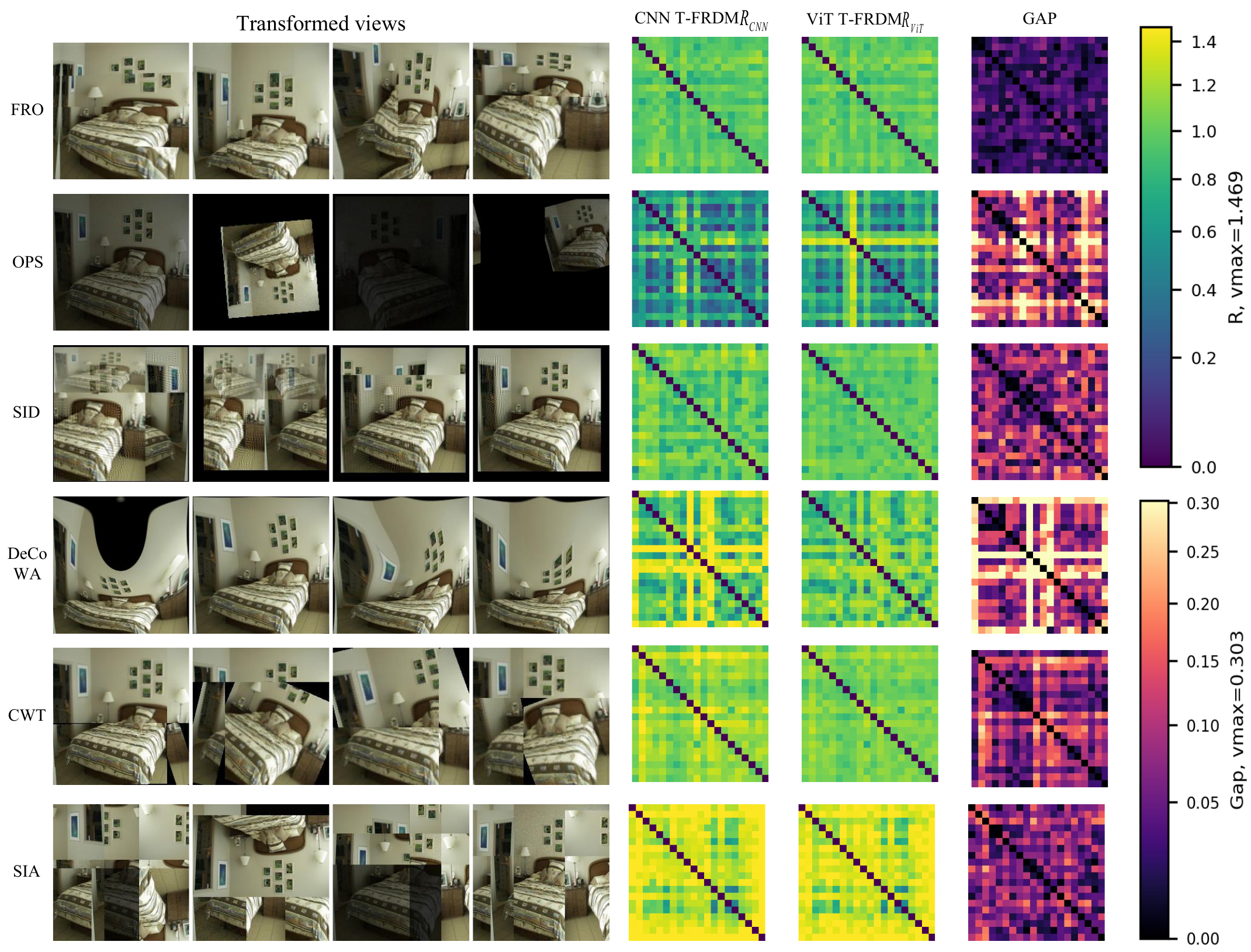}
    \caption{Qualitative comparison of transformed views and transformation-induced frontend response relations for different methods on a representative ImageNet image. Each row corresponds to one transformation method, while the columns show representative transformed views, the CNN T-FRDM, the ViT T-FRDM, and their element-wise absolute gap, respectively. All CNN and ViT T-FRDMs share the same color scale, whereas all GAP maps share another common color scale. Darker values in the GAP column indicate smaller response-relation discrepancies between the CNN and ViT frontends.}
    \label{fig:real_gap_all}
\end{figure*}

\section{Efficiency: Runtime and GPU Memory}
\label{sec:efficiency}

We evaluate practical efficiency using average \emph{time per image} and \emph{peak GPU memory} under different sampling budgets \(N\). The results are summarized in \cref{tab:time_memory_byN}.

\begin{table*}[tb]
	\centering
	\caption{Runtime (seconds per image) and peak GPU memory (MB) under different sampling budgets $N$.}
	\label{tab:time_memory_byN}

	\begin{tabular}{lcccccccccccc}
		\toprule
		& \multicolumn{2}{c}{$N=10$} & \multicolumn{2}{c}{$N=17$} & \multicolumn{2}{c}{$N=21$} & \multicolumn{2}{c}{$N=26$} &
		\multicolumn{2}{c}{$N=37$} & \multicolumn{2}{c}{$N=101$} \\
		\cmidrule(lr){2-3}\cmidrule(lr){4-5}\cmidrule(lr){6-7}\cmidrule(lr){8-9}\cmidrule(lr){10-11}\cmidrule(lr){12-13}
		Method
		& Time & Mem & Time & Mem & Time & Mem & Time & Mem & Time & Mem & Time & Mem \\
		\midrule
		SIA    & 0.2632 & 2200 & 0.3704 & 2530 & 0.4320 & 2678 & 0.5405 & 2822 & 0.7692 & 3500 & 2.11 & 5662 \\
		DeCoWA & 2.42   & 2154 & 3.92   & 2154 & 4.724  & 1844 & 6.10   & 2154 & 8.56   & 2154 & 24.0 & 2154 \\
		OPS(10,5,4)    & 0.9615 & 1860 & 1.60   & 1880 & 2.022  & 1884 & 2.56   & 1886 & 3.34   & 1874 & 9.50 & 1892 \\
		CWT    & 0.3509 & 2214 & 0.5556 & 2544 & 0.6667 & 2706 & 0.7692 & 2846 & 1.09   & 3532 & 3.08 & 5740 \\
        SID    & 3.73 & 1856 & 6.49 & 1866 & 8.23 & 1866 & 10.19 & 1868 & 14.28   & 1868 & 39.52 & 1868 \\
		FRO   & 0.2128 & 2198 & 0.3226 & 2530 & 0.3860 & 2678 & 0.4505 & 2822 & 0.6667 & 3504 & 1.85 & 5660 \\
		\bottomrule
	\end{tabular}
\end{table*}

As \(N\) increases, the runtime of all methods increases because more transformed views require additional forward and backward evaluations. FRO consistently achieves the shortest runtime across all sampling budgets. At \(N=101\), FRO requires 1.85 seconds per image, compared with 2.11 seconds for SIA, 3.08 seconds for CWT, and substantially longer runtimes of 9.50, 24.0, and 39.52 seconds for OPS, DeCoWA, and SID, respectively. This result indicates that FRO introduces limited additional computational overhead and remains efficient even under a relatively large sampling budget.

Peak GPU memory is strongly influenced by the implementation strategy and does not necessarily increase proportionally with runtime. FRO, SIA, and CWT process transformed views in parallel, resulting in increased memory consumption as \(N\) grows. In contrast, OPS, DeCoWA, and SID maintain nearly constant peak memory because their released implementations process transformed views largely in a serial manner, thereby trading memory efficiency for considerably longer runtime. Overall, FRO provides the most favorable runtime performance while maintaining memory consumption comparable to other batch-parallel transformation methods.

\end{document}